# A Survey on Latent Tree Models and Applications

**Raphaël Mourad**　　　　　　　　　　　　　　　　　　　　　raphael.mourad@aliceadsl.fr
*LINA*, UMR CNRS *6241,*
*Ecole Polytechnique de l'Université de Nantes*
*Nantes, Cedex 3, 44306 France*

**Christine Sinoquet**　　　　　　　　　　　　　　　　　　　christine.sinoquet@univ-nantes.fr
*LINA*, UMR CNRS *6241, Université de Nantes*
*Nantes, Cedex, 44322 France*

**Nevin L. Zhang**　　　　　　　　　　　　　　　　　　　　　　　　　lzhang@cse.ust.hk
**Tengfei Liu**　　　　　　　　　　　　　　　　　　　　　　　　　　　liutf@cse.ust.hk
*Department of Computer Science & Engineering, HKUST*
*Clear Water Bay Road, Kowloon, Hong Kong*

**Philippe Leray**　　　　　　　　　　　　　　　　　　　　　philippe.leray@univ-nantes.fr
*LINA*, UMR CNRS *6241,*
*Ecole Polytechnique de l'Université de Nantes*
*Nantes, Cedex 3, 44306 France*

## Abstract

In data analysis, latent variables play a central role because they help provide powerful insights into a wide variety of phenomena, ranging from biological to human sciences. The latent tree model, a particular type of probabilistic graphical models, deserves attention. Its simple structure - a tree - allows simple and efficient inference, while its latent variables capture complex relationships. In the past decade, the latent tree model has been subject to significant theoretical and methodological developments. In this review, we propose a comprehensive study of this model. First we summarize key ideas underlying the model. Second we explain how it can be efficiently learned from data. Third we illustrate its use within three types of applications: latent structure discovery, multidimensional clustering, and probabilistic inference. Finally, we conclude and give promising directions for future researches in this field.

## 1. Introduction

In statistics, **latent variables** (LVs), as opposed to **observed variables** (OVs), are random variables which are not directly measured. A wide range of statistical models, called **latent variable models**, relate a set of OVs to a set of LVs. In these models, LVs explain dependences among OVs and hence offer compact and intelligible insights of data. Moreover LVs allow to reduce data dimensionality and to generate conditionally independent variables, which considerably simplifies downstream analysis. Applications are numerous and cover many scientific fields. This is typically the case in domains such as psychology, sociology, economics, but also biological sciences and artificial intelligence, to cite some examples. Such fields may need complex constructs that cannot be observed directly. For instance, human personality in psychology and social class in socio-economics refer to higher level abstractions than observed reality.

### 1.1 Context

**Latent tree model** (LTM)[1] is a class of latent variable models which has received considerable attention. LTM is a probabilistic tree-structured graphical model where leaf nodes are observed while internal nodes can be either observed or latent. This model is appealing since its simple structure - a tree - allows simple and efficient inference, while its latent variables capture complex relationships.

A subclass of LTMs was first developed in the phylogenetic community (Felsenstein, 2003). In this context, leaf nodes are observed taxa while internal nodes represent unobserved taxum ancestors. For instance, in molecular genetic evolution, which is the process of evolution at DNA scale, it is obviously hopeless to study DNA sequences in some dead species from collecting their DNA. Nevertheless, evolutionary latent models can be used to infer the most probable ancestral sequences knowing contemporary living species sequences. Neighbor joining represents one of the first algorithms developed for LTM learning in the phylogenetic context (Saitou & Nei, 1987; Gascuel & Steel, 2006). It is still very popular as it is quick to compute and it allows to find the optimal model in polynomial time under certain assumptions.

During the past decade, LTMs in their general form have been under extensive investigation and have been applied to many fields. For instance they have been applied in human interaction recognition. Human interaction recognition is a challenging task, because of multiple body parts and concomitant inclusions (Aggarwal & Cai, 1999). For this purpose, the use of LTM allows to segment the interaction in a multi-level fashion (Park & Aggarwal, 2003): body part positions are estimated through low-level LVs, while overall body position is estimated by a high-level LV. LTMs have also been used in medical diagnosis (Zhang, Yuan, Chen, & Wang, 2008). In this context, LTMs provide a way to identify, through the LVs, the different syndrome factors which cannot be directly observed by the physician.

### 1.2 Contributions

In this paper, we present a comprehensive study of LTM and a broad-brush view of its recent theoretical and methodological developments. LTM must be paid attention because (i) it offers deep insights for latent structure discovery (Saitou & Nei, 1987), (ii) it can be applied to multidimensional clustering (Chen, Zhang, Liu, Poon, & Wang, 2012) and (iii) it allows efficient probabilistic inference (Wang, Zhang, & Chen, 2008). Somewhat surprisingly, no extensive review on this research area has been published.

In addition to the reviewing of the LTM research area, we also contribute an analysis and a perspective that advance our understanding of the subject. We establish a categorization of learning methods. We present generic learning algorithms implementing fundamental principles. These generic algorithms are partly different from those of the literature because they have been adapted to a broader context. Besides, the performances of all the algorithms of the literature are compared in the context of small, large and very large simulated and real datasets. Finally, we discuss future directions, such as the adaptation of LTM for continuous data.

---

1. LTM has been previously called "hierarchical latent class model" (Zhang, 2004), but this name has been discarded because the model does not inherently reveal a hierarchy.

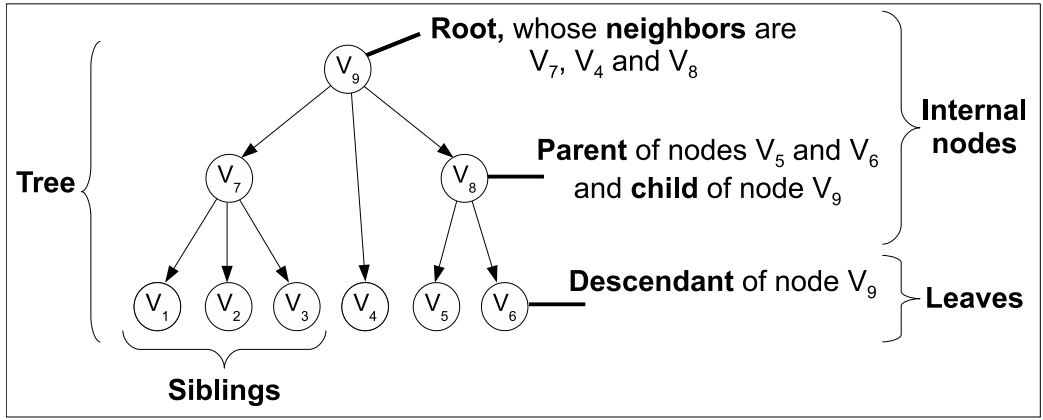

Figure 1: Illustration of graph theory terminology.

### 1.3 Paper Organization

This paper is organized as follows. Section 2 presents the latent tree model and related theoretical developments. In Section 3, we review methods developed to learn latent tree models for the two main situations: learning when structure is known and learning when it is not the case. Then, Section 4 presents and details three types of applications of latent tree models: latent structure discovery, multidimensional clustering and probabilistic inference. Other applications such as classification are also discussed. Finally, the last two sections 5 and 6 conclude and point out future directions.

## 2. Theory

In this section, we first introduce graph terminology and then present LTM. Latent classes and probabilistic inference for clustering are next presented. Scoring LTMs is discussed. We also present the concepts of marginal equivalence, equivalence and model parsimony, useful for LTM learning. Then, we explain the necessity of a trade-off between latent variable complexity and partial structure complexity.

Variables are denoted by capital letters, *e.g.* $A$, $B$ and $C$, whereas lower-case letters refer to values that variables can take, *e.g.* $a$, $b$ and $c$. Bold-face letters represent sets of objects, that is **A**, **B** and **C** are sets of variables while **a**, **b** and **c** are value sets. An observed variable is denoted $X$ whereas a latent variable is denoted $H$. A variable about which we do not know if it is observed or latent is denoted $V$.

### 2.1 Graph Theory Terminology

Before presenting LTM, we first need to define graph-related terms, which are illustrated in Figure 1. A graph $G(\mathbf{V}, \mathbf{E})$ is composed of a set of nodes **V** and a set of edges $\mathbf{E} \subset \mathbf{V} \times \mathbf{V}$. An edge is a pair of nodes $(V_a, V_b) \in \mathbf{E}$. The edge is undirected (noted $V_a - V_b$) if $(V_b, V_a) \in \mathbf{E}$ and directed (noted $V_a \rightarrow V_b$) if not.

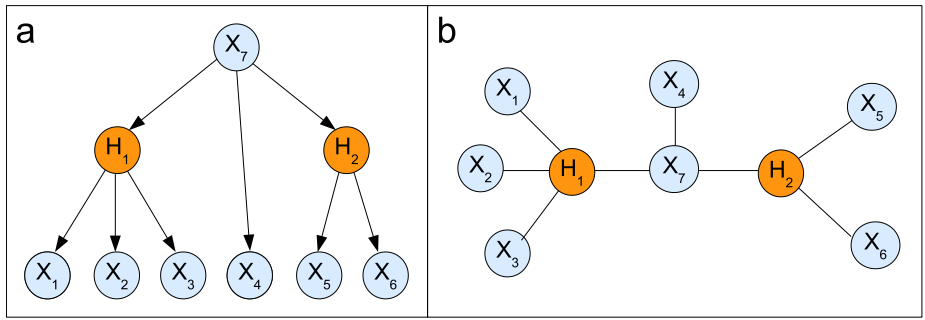

Figure 2: (a) Directed tree. (b) Undirected tree. The light shade (blue) indicates the observed variables whereas the dark shade (red) points out the latent variables.

A directed graph is a graph whose all edges are directed. In a directed graph, a node $V_a$ is a parent of a node $V_b$ if and only if there exists an edge from $V_a$ to $V_b$. The node $V_b$ is then called the child of node $V_a$. Nodes are siblings if they share the same parent. A node $V_c$ is a root if it has no parent. A directed path from a node $V_d$ to a node $V_a$ is a sequence of nodes such that for each node except the last one, there is an edge to the next node in the sequence. A node $V_a$ is a descendant of a node $V_d$ if and only if there is a directed path from $V_d$ to $V_a$. The node $V_d$ is then called an ancestor of node $V_a$.

An undirected graph only contains undirected edges. In an undirected graph, a node $V_a$ is a neighbor of another node $V_b$ if and only if there is an edge between them. A leaf is a node having only one neighbor. An internal node is a node having at least two neighbors. An undirected path is a path for which the edges are not all oriented in the same direction.

A clique is a set of pairwise connected nodes (in a tree, a clique is simply an edge). A separator is a set of nodes whose removal disconnect two or more cliques (in a tree, a separator is simply an internal node). A tree $T$ is a graph for which any two nodes are connected by exactly one path.

## 2.2 Latent Tree Model

LTM is a tree-structured graphical model with latent variables. It is composed of a tree - the **structure** - $T(\mathbf{V}, \mathbf{E})$, and a set of parameters, $\theta$. The tree can be either directed (*i.e.* a Bayesian network; Zhang, 2004) or undirected (*i.e.* a Markov random field; Choi, Tan, Anandkumar, and Willsky, 2011). Both representations are described in Figure 2. The set of nodes $\mathbf{V} = \{V_1, ..., V_{n+m}\}$ represents $n+m$ observed and latent variables. $\mathbf{X} = \{X_1, ..., X_n\}$ is the set of observed variables and $\mathbf{H} = \{H_1, ..., H_m\}$ is the set of latent variables. Leaf nodes are OVs while internal nodes can be either observed or latent. Variables can be either discrete or continuous. The set of $k$ edges $\mathbf{E} = \{E_1, ..., E_k\}$ captures the direct dependences between these variables.

In the directed setting (Figure 2a), the set of parameters $\theta$ consists of probability distributions, one for each variable. Given a variable $V_i$ with parents $Pa_{V_i}$, a conditional distribution $P(V_i|Pa_{V_i})$ is defined. For a variable $V_i$ that has no parent, a marginal distribution $P(V_i)$ is defined instead. The joint probability distribution (JPD) of this model is

formulated as:

$$P(\mathbf{V}) = \Pi_{i=1}^{n+m} P(V_i | Pa_{V_i}). \tag{1}$$

To illustrate the model, let us take the example in Figure 2a. It is composed of a set of OVs $\{X_1, ..., X_7\}$ and a set of LVs $\{H_1, H_2\}$. Its JPD writes as:

$$\begin{aligned} P(X_1, ..., X_7, H_1, H_2) &= P(X_1|H_1) \, P(X_2|H_1) \, P(X_3|H_1) \, P(H_1|X_7) \, P(X_7) \\ &\times P(X_4|X_7) \, P(H_2|X_7) \, P(X_5|H_2) \, P(X_6|H_2). \end{aligned} \tag{2}$$

In the undirected setting (Figure 2b), the set of parameters $\theta$ consists of probability distributions, one for each clique and separator. In LTM, cliques are edges while separators are internal nodes. Let $\{I_1, ..., I_j\}$ be the separators. The JPD of this model is formulated as:

$$P(\mathbf{V}) = \frac{\Pi_{(V_a, V_b) \in \mathbf{E}} \, P(V_a, V_b)}{\Pi_{j=1}^{m} \, P(I_j)^{(d(I_j)-1)}}, \tag{3}$$

where $d(I_j)$ is the degree of internal node $I_j$. The JPD of the undirected model in Figure 2b writes as:

$$\begin{aligned} P(X_1, ..., X_7, H_1, H_2) &= \frac{P(X_1, H_1) \, P(X_2, H_1) \, P(X_3, H_1) \, P(H_1, X_7) \, P(X_4, X_7)}{P(H_1) \, P(H_1) \, P(H_1) \, P(X_7)} \\ &\times \frac{P(X_7, H_2) \, P(X_5, H_2) \, P(X_6, H_2)}{P(X_7) \, P(H_2) \, P(H_2)}. \end{aligned} \tag{4}$$

In the following, for the seek of simplicity, we restrain the study to categorical variables, *i.e.* random variables which have a finite number of states. We also mainly focus on LTM whose internal nodes are all LVs. Most works on LTM have been developed for these two model settings.

### 2.3 Latent Classes and Clustering

An LV has a number of states, each of them representing a latent class. All latent classes together represent a soft partition of the data and define a finite mixture model (FMM). LTM can be seen as multiple FMMs connected to form a tree. Given a data point, the probability of belonging to a particular class can be computed using the Bayes formula. This computation is called class assignment.

In an LTM, each LV $H_j$ represents a partition of the data. For an observation $\ell$ and the vector of its values $\mathbf{x}^\ell = \{x_1^\ell, ..., x_n^\ell\}$ on the set of OVs $\mathbf{X}$, the probability of its membership to a class $c$ of an LV $H_j$ can be computed as follows:

$$\begin{aligned} P(H_j = c | \mathbf{x}^\ell) &= \frac{P(\mathbf{x}^\ell | H_j = c) \, P(H_j = c)}{P(\mathbf{x}^\ell)} \\ &= \frac{\sum_{\mathbf{H}'} P(\mathbf{x}^\ell, \mathbf{H}' | H_j = c) \, P(H_j = c)}{\sum_{c=1}^{k} \sum_{\mathbf{H}'} P(\mathbf{x}^\ell, \mathbf{H}' | H_j = c) \, P(H_j = c)} \end{aligned} \tag{5}$$

with $\mathbf{H}' = \mathbf{H} \backslash \{H_j\}$ and $k$ the cardinality of $H_j$. From the last formula, the reader might get the impression that the complexity of class assignment is exponential. However, in trees, linear[2] - thus efficient - probabilistic inference, using message passing (Kim & Pearl, 1983), can be used to compute $P(H_j = c|\mathbf{x}^\ell)$.

Probabilistic inference of LV values has two applications: clustering and latent data imputation. Clustering using LTMs will be illustrated and seen in detail in Section 4.2. Latent data imputation is the process of inferring, for each observation $\ell$, the values of LVs. These variables are called imputed LVs. In Section 3.2.2, we will see that latent data imputation is at the basis of fast variable clustering-based LTM learning methods.

### 2.4 Scoring Latent Tree Models

In theory, every score, such as Akaike information criterion (AIC) (Akaike, 1970) and **Bayesian information criterion** (BIC) (Schwartz, 1978), could be used for scoring LTMs. In practice, the BIC score is often used for LTMs. Let consider a set of $n$ OVs $\mathbf{X} = \{X_1, ..., X_n\}$ and a collection of $N$ identical and independently distributed (i.i.d.) observations $D_\mathbf{x} = \{\mathbf{x}^1, ..., \mathbf{x}^N\}$. BIC is composed of two terms:

$$BIC(T, D_\mathbf{x}) = log\ P(D_\mathbf{x}|\theta^{ML}, T) - \frac{1}{2}dim(T)\ log\ N, \qquad (6)$$

with $\theta^{ML}$ the maximum likelihood parameters, $dim(T)$ the model dimension and $N$ the number of observations. The first term evaluates the fit of the model to the data. It is computed through probabilistic inference of $P(\mathbf{x}^\ell)$ for each observation $\ell$. The second term of the score penalizes the model according to its dimension, to prevent overfitting.

In models without latent variables, the dimension is simply calculated as the number of free parameters. This is sometimes called standard dimension. When LVs are present, standard dimension is no longer an appropriate measure of model complexity, and effective dimension should be used instead (Geiger, Heckerman, & Meek, 1996). Effective dimension is computed as the rank of the Jacobian matrix of the mapping from the model parameters to the OV joint distribution.

### 2.5 Model Parsimony

Let us consider two LTMs, $\mathcal{M} = (T, \theta)$ and $\mathcal{M}' = (T', \theta')$, built on the same set of $n$ OVs, $\mathbf{X} = \{X_1, ..., X_n\}$. We say that $\mathcal{M}$ and $\mathcal{M}'$ are **marginally equivalent** if their joint distributions on OVs are equal:

$$P(X_1, ..., X_n|T, \theta) = P(X_1, ..., X_n|T', \theta'). \qquad (7)$$

If two marginally equivalent models have the same dimension, they are **equivalent** models.

A model $\mathcal{M}$ is **parsimonious**[3] if there does not exist another model $\mathcal{M}'$ that is marginally equivalent and has a smaller dimension. A parsimonious model has the best possible score. It does not contain any redundant LVs or any redundant latent classes. It represents the model to infer from data. Two conditions ensure that an LTM does not include any redundant LVs (Pearl, 1988):

---

2. Actually, in trees, inference is linear with the number of edges $|\mathbf{E}|$, and is thus also linear with the number $n + m$ of observed and latent variables, because $|\mathbf{E}| \leq n + m - 1$.
3. The notion of parsimony is also called minimality by Pearl (1988).

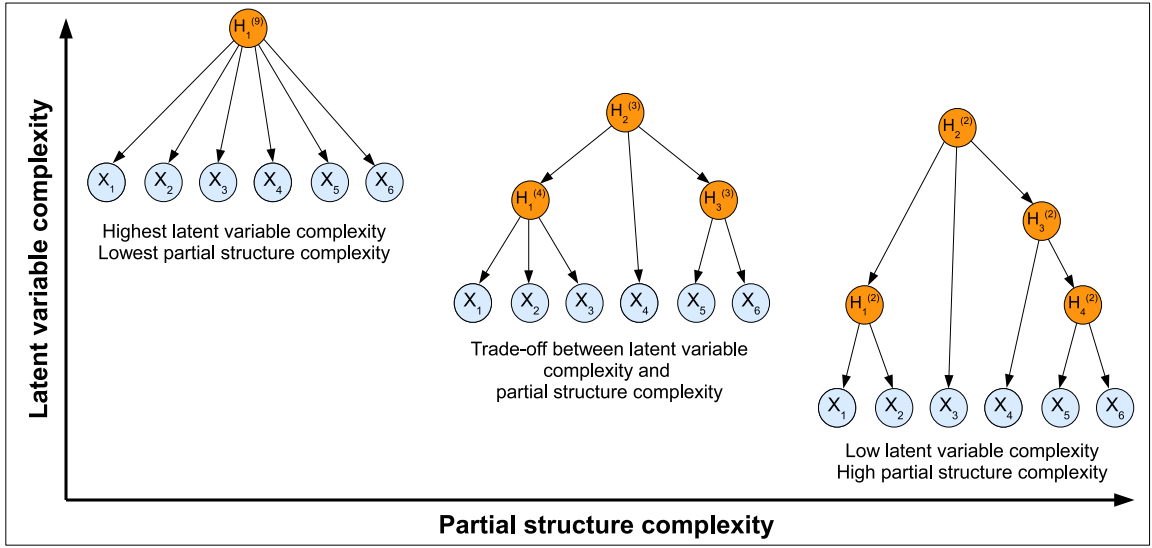

Figure 3: Illustration of the trade-off between latent variable complexity and partial structure complexity in latent tree models. Superscript represents LV cardinality. See Figure 2 for node color code.

- An LV must have at least three (observed or latent) neighbors. If it has only two neighbors, it can simply be replaced by a direct link between the two.

- Any two variables connected by an edge in the LTM are neither perfectly dependent nor independent.

There is also a condition ensuring that an LTM does not include any redundant latent classes. Let $H$ be an LV in an LTM. The set of $k$ variables $Z = \{Z_1, ..., Z_k\}$ are the neighbors of $H$. An LTM is **regular** (Zhang, 2004) if for any LV $H$:

$$|H| \leq \frac{\Pi_{i=1}^{k}|Z_i|}{\max_{i=1}^{k}|Z_i|}. \tag{8}$$

Zhang (2004) showed that all parsimonious models are necessarily regular. Thus the model search can be restricted to the space of regular models. Zhang also demonstrated that the space of regular models is upper bounded by $2^{3n^2}$, where $n$ is the number of OVs.

## 2.6 Trade-off between Latent Variable Complexity and Partial Structure Complexity

Zhang and Kocka (2004b) distinguished two kinds of model complexity in LTM: **latent variable complexity** refers to LV cardinalities while **partial structure complexity**[4] is the edges and number of LVs in the graph. The trade-off between these two complexities has an important role to play when one wants to choose a model. This trade-off is illustrated in Figure 3. For instance, let us consider a latent class model (*i.e.* a model with only one LV, abbreviated LCM) *versus* an LTM having the same marginal likelihood (marginally equivalent models). The LCM is the model showing the highest LV complexity and the lowest partial structure complexity. It might have a low score if local dependences are present between OVs. At the opposite, the model with a low LV complexity and a high partial structure complexity, a binary tree with binary LVs, would also have a low score, because some LVs would be unnecessary. Depending on the application, a model showing a good trade-off between the two complexities should be preferred, because it would present a better score and might be easier to interpret.

---

4. In their paper, Zhang and Kocka (2004b) called it structure complexity. For a better understanding, we prefer to make the distinction between (complete) structure which includes LV cardinalities and partial structure which does not.

## 3. Statistical Learning

In this section, we present generic algorithms implementing fundamental principles for learning LTMs. These algorithms are partly different from those proposed in the literature, because they have been adapted to a broader context. Moreover we provide a unified presentation of algorithms, in the context of this survey. When learning a model from data, two main situations have to be distinguished: when structure is known and only parameters have to be learned, and the more complicated situation where both are unknown.

### 3.1 Known Structure

In the simplest situation, structure is known, *i.e.* not only the dependences between variables but also the number of LVs and their respective cardinalities (*i.e.* the numbers of latent classes). The problem is to estimate probability parameters. To solve the problem, one can use expectation-maximization (EM), the most popular algorithm for learning parameters in the face of LVs (Dempster, Laird, & Rubin, 1977; Lauritzen, 1995). Because EM leads to computational burden for large LTMs, a more efficient procedure, that we call LCM-based EM, can be used. Other methods different from EM, such as spectral techniques, have also been developed.

#### 3.1.1 Expectation-maximization

Ideally, when learning parameters, we would like to maximize the log-likelihood for a set of $N$ i.i.d. observed data $D_{\mathbf{x}} = \{\mathbf{x}^1, ..., \mathbf{x}^N\}$:

$$L(\theta; D_{\mathbf{x}}) = \log\ P(D_{\mathbf{x}}|\theta) = \log \sum_{\mathbf{H}} P(D_{\mathbf{x}}, \mathbf{H}|\theta). \tag{9}$$

However, directly maximizing $L(\theta; D_{\mathbf{x}})$ in Equation (9) is often intractable because it involves the logarithm of a (large) sum. To overcome the difficulty, EM implements an iterative approach. At each iteration, it optimizes instead the following expected log-likelihood conditional on current parameters $\theta^t$:

$$Q(\theta; \theta^t) = E_{D_{\mathbf{h}}|D_{\mathbf{x}}, \theta^t}[\log\ P(D_{\mathbf{x}}, D_{\mathbf{h}}|\theta)] \tag{10}$$

where $D_{\mathbf{x}}$ is completed by the missing data $D_{\mathbf{h}} = \{\mathbf{h}^1, ..., \mathbf{h}^N\}$ inferred using $\theta^t$. Note that by completing the missing data, EM can easily deal with partially observed variables. An

---

**Algorithm 1** LCM-based EM parameter learning (LCMB-EM, adapted from Harmeling and Williams, 2011)

---

**INPUT:**
$T$, the tree structure of the LTM.

**OUTPUT:**
$\theta$, the parameters of the LTM.

1: $T' \leftarrow graph\_rooting(T)$ /* choose an LV as a root of $T$ */
2: $\mathbf{H_o} \leftarrow \emptyset$ /* initialization of the set of imputed latent variables */
3: $\theta' \leftarrow \emptyset$
4: **loop**
5:     $\mathbf{T_{LCM}} = \{T_{LCM_1}, ..., T_{LCM_k}\} \leftarrow identify\_LCMs\_of\_graph(T', \mathbf{H_o})$
6:     $\Theta_{\mathbf{LCM}} = \{\theta_{LCM_1}, ..., \theta_{LCM_k}\} \leftarrow EM(\mathbf{T_{LCM}})$
7:     **if** $|\mathbf{T_{LCM}}| > 1$ **then**
8:        $\theta' \leftarrow \theta' \cup children\_parameters(\Theta_{\mathbf{LCM}})$
9:     **else**
10:        $\theta' \leftarrow \theta' \cup children\_and\_parent\_parameters(\Theta_{\mathbf{LCM}})$
11:        **break**
12:     **end if**
13:     $\mathbf{H_o} \leftarrow \mathbf{H_o} \cup impute\_LV\_data(\Theta_{\mathbf{LCM}})$ /* now parents are observed */
14: **end loop**
15: $\theta \leftarrow EM(\theta')$ /* global EM using $\theta'$ as a starting point */

---

important drawback of EM is that it does not guarantee to reach the global optimum. To reduce the probability of getting trapped into a local maximum, random restarts (multiple starts with different random initial parameters) or simulated annealing represent well-used solutions. Wang and Zhang (2006) showed that a few random restarts suffice when the LTM is small and variables are strongly dependent with each other. In Supplemental material B.1, we also present our experiments on random restarts for EM. It appears that it is not possible to give a simple answer to how many restarts should be used because it depends on the model. Besides, convergence is sometimes not reached even after a very large number of restarts.

### 3.1.2 LCM-BASED EM

Although inference is linear in LTMs, running EM might be prohibitive. One solution to speed up EM computations consists in chaining two steps: a first step of divide-and-conquer strategy through local - LCM - learning, followed by a final step carrying out global learning. This LCM-based EM (**LCMB-EM**) parameter learning method[5] is presented in Algorithm 1 and illustrated in Figure 4. In the first step, parameters are locally learned through a bottom-up LCM-based learning procedure explained as follows. An LV is first chosen as a root of the LTM (line 1; note that this point will be discussed in detail in Section 3.2.5). In the rooted LTM, every LCM $\{\mathbf{X}, H\}$ is identified (line 5), *i.e.* every LV $H$ whose all children are OVs $\mathbf{X}$. Then, LCM parameters can be quickly learned through EM (line 6) and used to update current LTM parameters (lines 7 to 12). Once parameters have been learned for such an LCM, the distributions of unknown values of LV $H$ can be probabilistically inferred for each observation[6] (line 13; for more details, see Section 2.3, paragraph 2). These distributions can then be used as weighted observations. In their turn, these weighted observations will seed the learning processes for the LCMs on the (now observed) LVs $\mathbf{H_o}$ and remaining OVs which have not been used for LCM learning yet. Iterating the two operations (LCM learning and latent data imputation) leads to a bottom-up LCM-based EM procedure enabling local and fast parameter learning of LTMs. Finally, LTM parameters are refined through global EM using as starting point the locally learned parameters (line 15). As with EM, we also provide the results of experiments evaluating the efficiency of random restarts (see Supplemental material B.1). Our results show that LCMB-EM converges better than EM. However convergence is not achieved for the largest dataset studied.

---

5. This learning procedure is very similar to the one proposed for binary trees by Harmeling and Williams (2011).

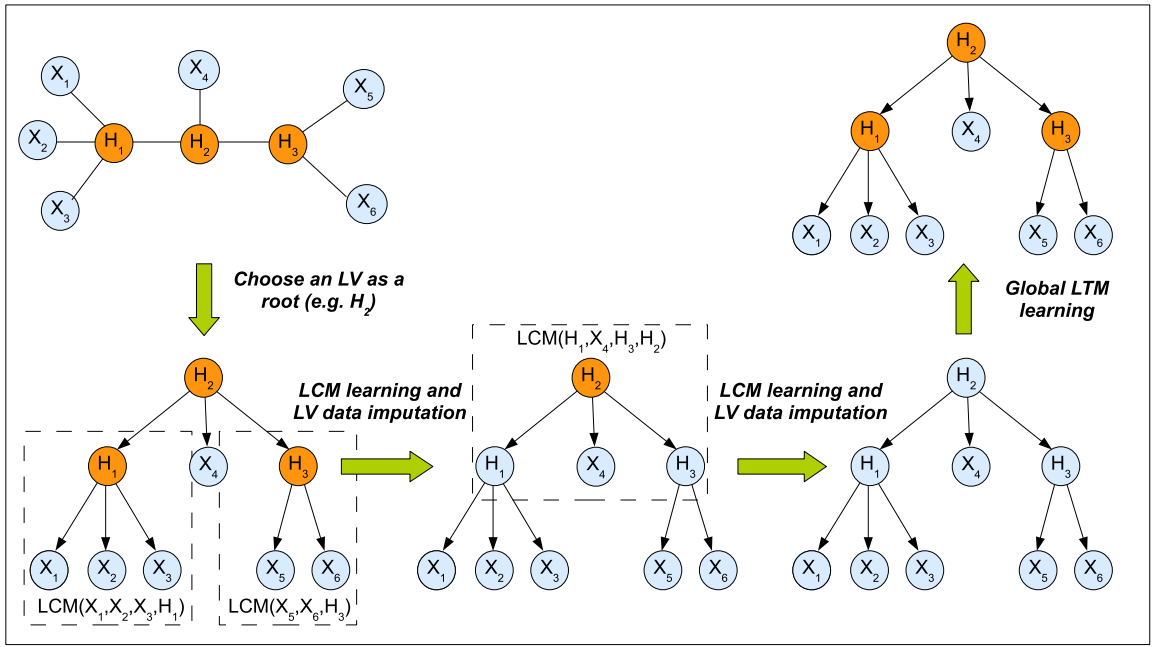

Figure 4: Illustration of LCM-based EM parameter learning algorithm (adapted from Harmeling and Williams, 2011). See Figure 2 for node color code.

### 3.1.3 Spectral Methods

Recently, Parikh et al. (2011) applied spectral techniques to LTM parameter learning. The method directly estimates the joint distribution of OVs without explicitly recovering the LTM parameters. Their algorithm is useful when LTM parameters are not required, for instance, for probabilistic inference over OVs only. The work of Parikh et al. alleviates the restriction of the approach of Mossel et al. (2006) requiring that all conditional probability tables should be invertible, and generalizes the method of Hsu et al. (2009) specific to hidden Markov models.

---
6. We recall that this process is named latent data imputation.

Parikh et al. (2011) reformulated the message passing algorithm using an algebraic formulation:

$$P(\mathbf{x}) = \sum_{\mathbf{H}} \Pi_{i=1}^{n+m} P(v_i | Pa_{v_i})$$
$$= \mathbf{r}^\top (\mathbf{M}_{j_1} \mathbf{M}_{j_2} ... \mathbf{M}_{j_J} \mathbf{1}_r), \tag{11}$$

where $\mathbf{x} = \{x_1, ..., x_n\}$ is an observation, $X_r$ is a root, $\mathbf{r}$ is the marginal probability vector of the root and $\mathbf{M}_{j_1}, \mathbf{M}_{j_2}, ..., \mathbf{M}_{j_J}$ are the incoming message matrices from the root's children. Children message matrices are calculated in a similar manner:

$$\mathbf{M}_i = \mathcal{T}_i \bar{\times}_1 (\mathbf{M}_{j_1} \mathbf{M}_{j_2} ... \mathbf{M}_{j_J} \mathbf{1}_i), \tag{12}$$

where $X_i$ is a root child, $\mathbf{M}_i$ is the outgoing message matrix from $X_i$, $\mathcal{T}_i$ is a third order tensor related to the conditional probability matrix between $X_i$ and $Pa_{X_i}$ (i.e. $X_r$), $\mathbf{M}_{j_1}, \mathbf{M}_{j_2}, ..., \mathbf{M}_{j_J}$ are the $X_i$'s children incoming messages and $\bar{\times}_1$ is the mode-1 vector product. Such as in the original message passing algorithm, all messages are recursively calculated starting from the leaves and going up to the root.

The drawback of the previous representation is that message passing still needs model parameters. To tackle this issue, the key is to recover $P(\mathbf{x})$ using invertible transformations. Message matrices are then calculated by transforming each message $\mathbf{M}_j$ by two invertible matrices $\mathbf{L}_j$ and $\mathbf{R}_j$ ($\mathbf{L}_j \mathbf{R}_j^{-1} = \mathbf{I}$):

$$\mathbf{M}_i = \mathcal{T}_i \bar{\times}_1 (\mathbf{L}_{j_1} \mathbf{L}_{j_1}^{-1} \mathbf{M}_{j_1} \mathbf{R}_{j_1} \mathbf{L}_{j_2}^{-1} \mathbf{M}_{j_2} \mathbf{R}_{j_2} ... \mathbf{L}_{j_J}^{-1} \mathbf{M}_{j_J} \mathbf{R}_{j_J} \mathbf{R}_{j_J}^{-1} \mathbf{1}_i). \tag{13}$$

The matrices $\mathbf{L}_j$, $\mathbf{M}_j$ and $\mathbf{R}_j$ can be recovered from singular vectors $\mathbf{U}_j$ of the empirical probability matrices $P(X_{\lambda_j}, X_j)$ of OVs $X_j$ and their left neighbor OV $X_{\lambda_j}$. This leads to a very efficient computation of message passing only involving a sequence of singular value decompositions of empirical pairwise joint probability matrices. We refer to the work of Parikh et al. (2011) for more details about the singular value decomposition and spectral algorithm. Compared to EM, this spectral method does not entail the problem of getting trapped in local maxima. Moreover, it performs comparable to or better than EM while being orders of magnitude faster.

### 3.1.4 Other Methods

Other methods exist for parameter learning. Gradient descent (Kwoh & Gillies, 1996; Binder, Koller, Russel, & Kanazawa, 1997) and variations of the Gauss-Newton method (Xu & Jordan, 1996) help accelerate the sometimes slow convergence of EM. However, they require the evaluation of first and/or second derivatives of the likelihood function. For a Bayesian learning, variational Bayes (Attias, 1999) offers a counterpart of EM.

### 3.2 Unknown Structure

Regrettably, most of the time, there is no *a priori* information on the LTM structure. This compels to learn every part of the model, *i.e.* the number of LVs, their cardinalities, the dependences and the parameters. This learning task represents a challenging issue, for which various methods have been conceived. In this section, we provide a survey of those algorithms. The determination of LV cardinalities, as well as the time complexity and scalability of algorithms are also discussed. We end by establishing a summary relative to these learning methods.

Structure learning approaches fall into three categories. The first one is comprised of search-based methods, inspired from standard Bayesian network learning. The second one is based on variable clustering and is related to hierarchical procedures. The last category relies on the notion of distances and comes from phylogenetics.

#### 3.2.1 Search-based Methods

Search-based methods aim at finding the model that is optimal according to some scoring metric. For BNs without LVs, the BIC score is often used. In the context of LTM, BIC suffers from a theoretical shortcoming as pointed out in Section 2.4. However, empirical results indicate that the shortcoming does not seem to compromise model quality in practice (Zhang & Kocka, 2004a). So, researchers still use BIC when it comes to learning LTM. Many search procedures have been proposed. They all explore the space of regular LTMs. Here we focus on: (i) the most naive one which is conceptually simple but very computationally expensive and (ii) the most advanced one which reduces the search space and implements fast parameter learning through local EM.

**Naive Greedy Search**

Naive greedy search (**NGS**) consists in starting from an LCM and then visiting the space of regular LTM partial structures. The neighborhood of the current model is explored by greedy search through operations such as addition or removal of a latent node, and node relocation[7]. For each partial structure neighbor, the cardinalities of all LVs are optimized through addition or dismissal of a state relative to an LV. During the model search (partial structure and LV cardinality), candidate models are learned with EM and evaluated through a score. If the best candidate model shows a score superior to the current model score, then the former is used as a seed for the next step. Otherwise, NGS stops and the current model is considered as the best model. Therefore, at each step of the search, the learning approach necessitates to evaluate the score of a very large number of candidate models. This leads to a huge computational burden because, for each candidate model evaluation, the likelihood has to be computed through EM.

---

7. Node relocation picks a child of an LV and then grafts it as a child of another LV which is connected to the former LV.

**Advanced Greedy Search**

Advanced greedy search (**AGS**) relies on three strategies to reduce the complexity. Advanced greedy search is presented in Algorithm 2. First, AGS focuses on a smaller space of models to explore than NGS (Zhang & Kocka, 2004b). The algorithm performs partial structure search and LV cardinality exploration simultaneously. For this purpose, two additional operators are used: addition and removal of a latent state for an LV.

Second, AGS follows a grow-restructure-thin strategy to reduce again the complexity of the search space (Chen, Zhang, & Wang, 2008; Chen et al., 2012). The strategy consists in dividing the five operators into three groups. Each group is applied at a given step of the model search. First, latent node and latent state introduction (NI and SI, respectively) are used to make the current model more complex[8] (grow, line 3). Then, node relocation (NR) rearranges connections between variables (restructure, line 4). Finally, latent node and latent state deletion (ND and SD, respectively) make the current model simpler (thin, line 5).

Third, one needs to assess the BIC score of candidate models. Learning parameters of the new models is thus required. To achieve fast learning, Chen et al. (2008, 2012) do not compute likelihood but instead the so-called restricted likelihood through the local EM procedure (line 12). The principle relies only on optimizing parameters of variables whose connection or cardinality was changed in the candidate model. Parameters of remaining variables are kept identical as in the current model.

**Operation Granularity**

When starting from the simplest solution (an LCM), Zhang and Kocka (2004b) observed that the comparison of the BIC scores between the candidate model $T'$ and the current one $T$ might not be a relevant criterion. The problem is that this strategy always leads to increase the cardinality of the LCM, without introducing LVs in the model (see trade-off between LV complexity and partial structure complexity in Section 2.6). To tackle this issue, they propose instead to assess the so-called improvement ratio during the grow step:

$$IR^{BIC}(T', T|D_\mathbf{X}) = \frac{BIC(T', D_\mathbf{X}) - BIC(T, D_\mathbf{X})}{dim(T') - dim(T)}, \qquad (14)$$

that is the difference of the BIC scores between candidate model $T'$ and current model $T$ divided by the difference of their respective dimensions.

### 3.2.2 Methods Based on Variable Clustering

The major drawback of search-based methods is that the evaluation of maximum likelihood in presence of LVs, as well as the large space to explore through local search, still entails computational burden. Approaches relying on variable clustering represent efficient and much faster alternatives. All of them rely on two key points: grouping variables to identify LVs and constructing model through a bottom-up strategy. Three main categories have been developed, depending on the structures learned: binary trees, non-binary trees and

---

8. Note that node relocation is used locally after each NI to increase the number of its children.

**Algorithm 2** Advanced greedy search for LTM learning (AGS, adapted from EAST, Chen et al., 2012)

**INPUT:**
  $\mathbf{X}$, a set of $n$ observed variables $\{X_1, ..., X_n\}$.

**OUTPUT:**
  $T$ and $\theta$, respectively the tree structure and the parameters of the LTM constructed.

1: $(T^0, \theta^0) \leftarrow latent\_class\_model(\mathbf{X})$ /* LCM learning using EM */
2: **loop for i = 0, 1,... until convergence**
3:     $(T^{i'}, \theta^{i'}) \leftarrow local\_search(NI \cup SI, T^i, \theta^i)$ /* grow */
4:     $(T^{i''}, \theta^{i''}) \leftarrow local\_search(NR, T^{i'}, \theta^{i'})$ /* restructure */
5:     $(T^{i+1}, \theta^{i+1}) \leftarrow local\_search(ND \cup SD, T^{i''}, \theta^{i''})$ /* thin */
6: **end loop**
7:
8: /* description of the function $local\_search(operators, T, \theta)$ */
9: $(T^0, \theta^0) \leftarrow (T, \theta)$
10: **loop for j = 0, 1,... until convergence**
11:     $\mathbf{T} = \{T_1, ..., T_\ell\} \leftarrow neighborhood(operators, T^j) \bigcup T^j$
12:     $\Theta = \{\theta_1, ..., \theta_\ell\} \leftarrow local\_EM(\mathbf{T}, \theta^j)$ /* except for $T^j$ */
13:     $T^{j+1} \leftarrow \arg\max_{\mathbf{T}} \begin{cases} BIC(\mathbf{T}, \Theta)/dim(\mathbf{T}, \Theta) & \textbf{if } operators = NI \cup SI \\ BIC(\mathbf{T}, \Theta) & \textbf{otherwise} \end{cases}$
14:     $\theta^{j+1} \leftarrow EM(T^{j+1})$
15:     **if** $T^{j+1} \in neighborhood(NI, T^j)$
16:       $(T^{j+1}, \theta^{j+1}) \leftarrow local\_search(NR, T^{j+1}, \theta^{j+1})$
17:     **end if**
18: **end loop**

forests.

### Binary Trees

Binary tree learning represents the most simple situation. For instance, one can simply compute an agglomerative hierarchical clustering (Xu & Wunsch, 2005) to learn the LTM structure. This algorithm is called agglomerative hierarchical cluster-based learning (**AHCB**). For the purpose of clustering, pairwise mutual information (MI) represents a well-suited similarity measure between variables (Cover & Thomas, 1991; Kraskov & Grassberger, 2009). The MI between two variables $V_i$ and $V_j$ can be defined as follows:

$$I(V_i; V_j) = \sum_{v_i \in V_i} \sum_{v_j \in V_j} p(v_i, v_j) \, \log \, \frac{p(v_i, v_j)}{p(v_i) \, p(v_j)}. \tag{15}$$

Single, complete or average linkage is used, depending on the cluster compactness required. The inferred hierarchy provides a binary tree (the partial structure) where leaf nodes are OVs and internal nodes are LVs (Wang et al., 2008; Harmeling & Williams, 2011). Then, LV cardinalities and parameters are learned (for details, see Section 3.2.4 and Section 3.1, respectively).

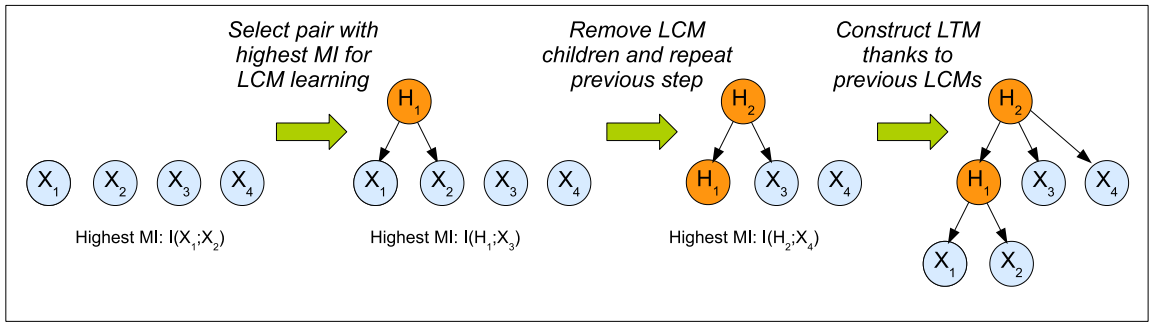

Figure 5: Illustration of the LCM-based LTM learning procedure for a set of 4 variables $\{X_1, X_2, X_3, X_4\}$.

In AHCB, each cluster of the hierarchy represents an LV. The MI between an LV and the other variables (observed or latent) is approximated through a linkage criterion. Instead of this approximation, another solution consists in directly computing the MI between variables, whatever their status, *i.e.* observed or latent (Hwang, Kim, & Zhang, 2006; Harmeling & Williams, 2011). To compute MI, values of LVs are imputed through LCM-based learning. Algorithm 3 presents the LCM-based LTM learning (**LCMB-LTM**) algorithm[9] implementing this solution. First, the working node set $\mathbf{W}$ is initialized with the set of OVs $\mathbf{X} = \{X_1, ..., X_n\}$ (line 1). An empty graph is created on $\mathbf{W}$ (line 2). Then the pair of variables showing the highest MI, $\{W_i, W_j\}$, is selected (line 5). An LCM ($\{W_i, W_j\}, H$) is learned (line 6), allowing to locally estimate the LV cardinality (through greedy search) and parameters, and then to impute values for $H$ (line 7). Once values of $H$ are known, it can be used as an observed variable (line 8). A new step of clustering, followed by LCM learning and LV value imputation, can then be performed. Iterating this step into a loop allows to construct an LTM through a bottom-up recursive procedure. At each step of the procedure, the parameters $\theta_{lcm}$ of the current LCM are used to update the current parameters $\theta'$ of the LTM (line 10). If only two nodes are remaining, the two nodes are connected[10] (line 12), the corresponding parameters are learned using maximum likelihood (line 13) and the loop is broken. After constructing the LTM, a final step globally learns LTM parameters using $\theta'$ as a starting point (line 17). LCMB-LTM yields slightly better BIC results than AHCB for large datasets (Harmeling & Williams, 2011). LCMB-LTM is illustrated for a set of 4 variables $\{X_1, X_2, X_3, X_4\}$ in Figure 5.

Harmeling and Williams (2011) justified selecting the pair of variables showing the highest MI at each step of LCMB-LTM. Let us consider a working node set of (observed or imputed latent) variables $\mathbf{W} = \{W_1, ..., W_\ell\}$. At each step of LCMB-LTM, the unknown

---

9. This algorithm is called LCMB-LTM to distinguish it from LCMB-EM for parameter learning (Algorithm 1). Both algorithms are similar and rely on LCM-based learning.
10. It prevents the introduction of a redundant LV, see Section 2.5, second paragraph.

**Algorithm 3** LCM-based LTM learning (LCMB-LTM, adapted from BIN-T, Harmeling and Williams, 2011)

**INPUT:**
$\mathbf{X}$, a set of $n$ observed variables $\{X_1, ..., X_n\}$.

**OUTPUT:**
$T(\mathbf{V}, \mathbf{E})$ and $\theta$, respectively the tree structure and the parameters of the LTM constructed.

1: $\mathbf{W} \leftarrow \mathbf{X}$ /* Initialization of the working set of variables */
2: $T(\mathbf{V}, \mathbf{E}) \leftarrow empty\_tree(\mathbf{W})$ /* tree on $\mathbf{W}$ with no edges */
3: $\theta' \leftarrow \emptyset$
4: **loop**
5:     $\{W_i, W_j\} \leftarrow pair\_with\_highest\_MI(\mathbf{W})$
6:     $lcm \leftarrow latent\_class\_model(\{W_i, W_j\})$
7:     $H \leftarrow impute\_LV\_data(lcm)$
8:     $\mathbf{W} \leftarrow \mathbf{W} \setminus \{W_i, W_j\} \cup H$ /* remove children and add imputed parent */
9:     $\mathbf{E} \leftarrow \mathbf{E} \cup edges(lcm); \mathbf{V} \leftarrow \mathbf{V} \cup H$
10:     $\theta' \leftarrow \theta' \cup children\_parameters(\theta_{lcm})$
11:     **if** $|\mathbf{W}| = 2$ **then**
12:         $\mathbf{E} \leftarrow \mathbf{E} \cup edge\_between\_the\_two\_remaining\_nodes(\mathbf{W})$
13:         $\theta' \leftarrow \theta' \cup learn\_remaining\_parameters(\mathbf{W})$ /* max likelihood estimation */
14:         **break**
15:     **end if**
16: **end loop**
17: $\theta \leftarrow EM(\theta')$ /* global EM using $\theta'$ as a starting point */

JPD $P(\mathbf{W})$ is approximated by a JPD $Q(\mathbf{W})$:

$$Q(\mathbf{W}) = P(W_i, W_j) \, \Pi_{W_k \in \mathbf{W} \setminus \{W_i, W_j\}} \, P(W_k)$$
$$= \frac{P(W_i, W_j)}{P(W_i) \, P(W_j)} \Pi_{W_k \in \mathbf{w}} \, P(W_k), \quad (16)$$

where only $W_i$ and $W_j$ are dependent. A proper measure to assess the divergence between $P(\mathbf{W})$ and $Q(\mathbf{W})$ is the Kullback-Leibler (KL) divergence, which is easy to calculate in this situation:

$$KL(P||Q) = \sum_{\mathbf{W}} P(\mathbf{W}) \, \log \, P(\mathbf{W}) - \sum_{\mathbf{W}} P(\mathbf{W}) \, \log \, Q(\mathbf{W})$$
$$= -I(W_i; W_j) + \sum_{\mathbf{W}} \, P(\mathbf{W}) \log \, \frac{P(\mathbf{W})}{\Pi_{W_k \in \mathbf{W}} P(W_k)}, \quad (17)$$

where $I(W_i; W_j)$ is the mutual information between $W_i$ and $W_j$. As the last term is constant, the maximization of the KL between $P$ and $Q$ simply consists in selecting the pair of variables with the highest MI before introducing an LV into the model under construction.

**Non-binary Trees**

Although modeling with binary trees performs well in practice (Harmeling & Williams, 2011), it would be worth alleviating the binarity restriction. Indeed it might provide a better model faithfulness and interpretation because less LVs would be required. There are several ways to learn non-binary trees without necessitating too much additional computational cost. For instance, Wang et al. (2008) first learn a binary tree. Then, they check each pair of neighbor LVs in the tree. If the information is redundant between the two LVs (*i.e.* the model is not parsimonious), then the LV node which is the child of the other LV node is removed and the remaining node is connected to every child of the removed node. Although the approach of Wang et al. is rigorous, it can lead in practice to find trees which are very close to binary trees.

Another solution consists in identifying cliques of pairwise dependent variables to detect the presence of LVs (Martin & Vanlehn, 1995; Mourad, Sinoquet, & Leray, 2011). For this purpose, Mourad et al. propose to alternate two main steps: (i) at each agglomerative step, a clique partitioning method is used to identify disjoint cliques of variables; (ii) each such clique, containing at least two nodes, is connected to an LV through an LCM. For each LCM, parameters are learned using EM and LV data are imputed through probabilistic inference. In Algorithm 3 (line 5), it is possible to replace the selection of the pair of variables having the highest MI by a clique partitioning step where each clique leads to construct an LCM and to impute corresponding LV values.

**Flat Trees**

All the algorithms discussed so far in this section are based on the idea of hierarchical variable clustering. The Bridged Island (BI) algorithm by Liu et al. (2012) takes a slightly different approach. It first partitions the set of all observed variables into subsets that are called sibling clusters. Then it creates an LCM for each sibling cluster by introducing a latent variable and optimizing its cardinality as well as the model parameters. After that, it imputes the values of the latent variables and links the latent variables up to form a tree structure using Chow-Liu's algorithm. The EM algorithm is run once at the end to optimize the parameters of the whole model. To highlight the difference between BI and the other variable clustering algorithms, we call the models it produces flat LTMs. Sibling cluster determination is the key step in BI. BI determines the sibling clusters one by one. To determine the first sibling cluster, it starts with the pair of variables with the maximum empirical MI. The cluster is expanded by adding other variables one after another. At each step, the variable that is the most dependent with the variables already in the cluster is added to the cluster. After that, a unidimensionality test (UD test) determines whether the dependences among all the variables in the cluster can be properly modeled with one latent variable. If the test fails, the expansion process is terminated and the first sibling cluster is determined. Thereafter, the same process repeats on the remaining observed variables until they are all grouped into sibling clusters.

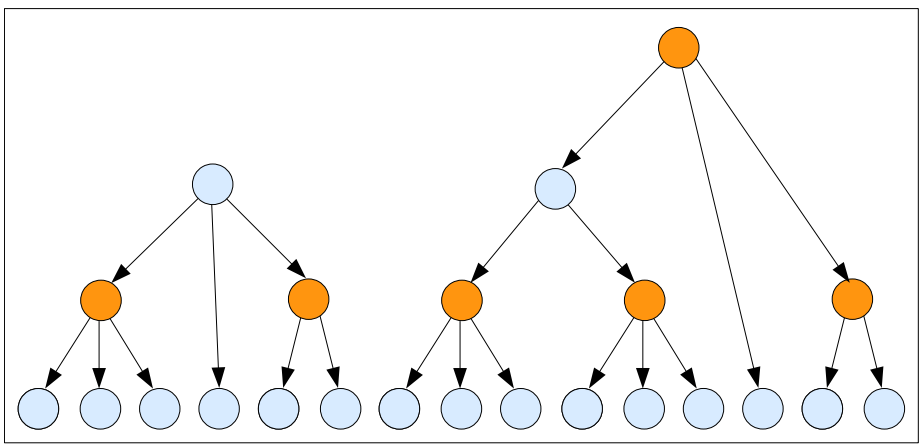

Figure 6: Latent forest models. See Figure 2 for node color code.

**Forests**

When the number of variables to analyze is very large (*e.g.* 1000 variables), it might be more reasonable to learn a forest instead of a tree because many variables might not be significantly dependent of each other (see Figure 6). We call this model "latent forest model" (LFM). It has many advantages over LTM, such as reducing the complexity of probabilistic inference (which depends on the number of edges). To learn LFM, there exists multiple approaches. For instance, in AHCB, one can use a cluster validation criterion to decide where to cut the hierarchy. Regarding LCMB-LTM (Algorithm 3), there are two options. On the one hand, Harmeling and Williams (2011) check the optimal cardinality of the current LV $H$ (additional step after line 6). If its optimal cardinality equals 1, this means that the LV is not useful to the model under construction and the algorithm stops. On the other hand, after partitioning variables in cliques (which replaces the step in line 5), the algorithm of Mourad et al. (2011) terminates when only single-node cliques are discovered.

To build an LFM, one can also first construct an LTM and then use the independence testing method of Tan et al. (2011) for pruning non-significant edges. This method provides guarantees to satisfy structural and risk consistencies. Similar works for non-parametric analysis have also been developed by Liu et al. (2011). It is worth mentioning that, to ensure model parsimony, the pruning of non-significant edges in an LTM should be followed by the removal of latent nodes which are no longer connected to a minimum of three nodes.

### 3.2.3 Distance-based Methods

This class of methods has been originally developed for phylogenetics (Felsenstein, 2003). A phylogenetic tree is a binary LTM showing the evolutionary relations among a set of taxa. Compared to the other LTM learning methods, distance-based ones provide strong guarantees for the inference of the optimal model. In this section, we first define distances and then present learning algorithms: neighbor joining (phylogenetic tree inference), distance-based

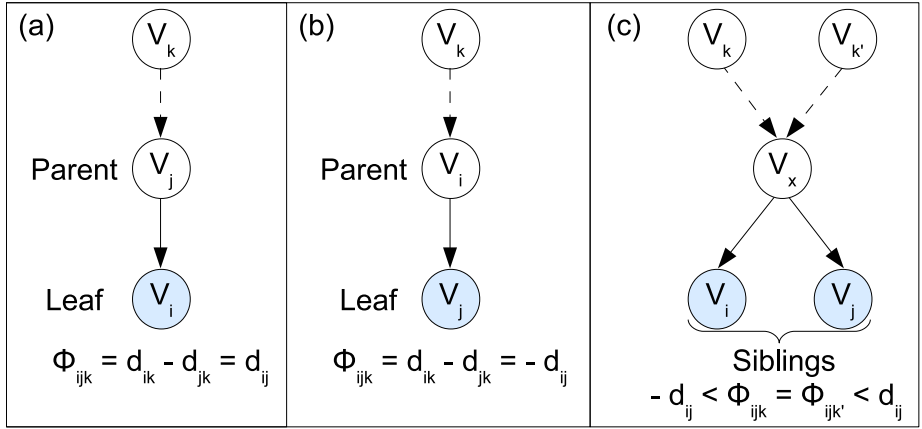

Figure 7: Illustration for ascertaining child-parent (a and b) and sibling relations (c). A dotted edge between two nodes means that the two nodes are linked by a path of unknown length. In this figure, some nodes are colored in white as they can be either observed or latent.

method dedicated to general LTM learning and recent spectral methods.

**Distances between Variables**

Distances are restricted to LTMs whose all variables share the same state space $\mathcal{X}$, *e.g.* binary variables (Lake, 1994). Distances are functions of pairwise distributions. For a discrete tree model $G(\mathbf{V}, \mathbf{E})$ (*e.g.* an LTM), the distance between two variables $V_i$ and $V_j$ is:

$$d_{ij} = -\log \frac{|\det(\mathbf{J}^{ij})|}{\sqrt{\det(\mathbf{M}^i)\ \det(\mathbf{M}^j)}}, \tag{18}$$

with $\mathbf{J}^{ij}$ the joint probability matrix between $V_i$ and $V_j$ (*i.e.* $J^{ij}_{ab} = p(V_i = a, V_j = b), a, b \in \mathcal{X}$), and $\mathbf{M}^i$ the diagonal marginal probability matrix of $V_i$ (*i.e.* $M^i_{aa} = p(V_i = a)$).

For a special case of discrete tree models, called symmetric discrete tree models (Choi et al., 2011), distance has a simpler form. Symmetric discrete tree models are characterized by the fact that every variable has a uniform marginal distribution and that any pair of variables $V_i$ and $V_j$ connected by an edge in $\mathbf{E}$ verifies the following property:

$$p(v_i|v_j) = \begin{cases} 1 - (K-1)\ \theta_{ij} & \text{if } v_i = v_j \\ \theta_{ij} & \text{otherwise}, \end{cases}$$

with $K$ the cardinality common to $V_i$ and $V_j$, and $\theta_{ij} \in (0, 1/K)$, known as the crossover probability. For symmetric discrete tree models, the distance between two variables $V_i$ and $V_j$ is then:

$$d_{ij} = -(K-1)\ \log\ (1 - K\theta_{ij}). \tag{19}$$

Note that there is a one-to-one correspondence between distances and model parameters for symmetric discrete tree models (for more details, see Choi et al., 2011).

The aforementioned distances are additive tree metrics (Erdos, Szekely, Steel, & Warnow, 1999):

$$d_{k\ell} = \sum_{(V_i, V_j) \in Path((k,\ell); \mathbf{E})} d_{ij}, \quad \forall k, \ell \in \mathbf{V}. \tag{20}$$

Choi et al. (2011) showed that additive tree distances allow to ascertain child-parent and sibling relationships between variables in a parsimonious LTM. Let us consider any three variables $V_i, V_j, V_k \in \mathbf{V}$. Choi et al. define $\Phi_{ijk}$ as the difference between distances $d_{ik}$ and $d_{jk}$ ($\Phi_{ijk} = d_{ik} - d_{jk}$). For any distance $d_{ij}$ between $V_i$ and $V_j$, the following two properties on $\Phi_{ijk}$ hold:

- $\Phi_{ijk} = d_{ij}, \forall V_k \in \mathbf{V} \setminus \{V_i, V_j\}$, if and only if $V_i$ is a leaf node and $V_j$ is its parent node;

- $\Phi_{ijk} = -d_{ij}, \forall V_k \in \mathbf{V} \setminus \{V_i, V_j\}$, if and only if $V_j$ is a leaf node and $V_i$ is its parent node;

- $-d_{ij} < \Phi_{ijk} = \Phi_{ijk'} < d_{ij}, \forall V_k, V_{k'} \in \mathbf{V} \setminus \{V_i, V_j\}$, if and only if $V_i$ and $V_j$ are leaf nodes and they share the same parent node, *i.e.* they belong to the same sibling group.

**Neighbor Joining**

The principle of neighbor joining (NJ) is quite simple (Saitou & Nei, 1987; Gascuel & Steel, 2006). NJ starts with a star-shaped tree. Then it iteratively selects two taxa $i$ and $j$, and it creates a new taxum $u$ to connect them. The selection of a pair seeks to optimize the following $\mathcal{Q}$ criterion:

$$\mathcal{Q}(i,j) = (n-2) \, d_{ij} - \sum_{k=1}^{n} d_{ik} - \sum_{k=1}^{n} d_{jk}, \tag{21}$$

where $n$ is the number of taxa and $d_{ij}$ is the additive tree distance between $i$ and $j$. The distance between $i$ and the new taxum $u$ is estimated as follows:

$$d_{iu} = \frac{1}{2} d_{ij} + \frac{1}{2(n-2)} \left( \sum_{k=1}^{n} d_{ik} - \sum_{k=1}^{n} d_{jk} \right), \tag{22}$$

and $d_{ju}$ is calculated by symmetry. The distances between the new taxum $u$ and the other taxa in the tree are computed as:

$$d_{uk} = \frac{1}{2} (d_{ik} - d_{iu}) + \frac{1}{2} (d_{jk} - d_{ju}). \tag{23}$$

The success of distance methods such as NJ comes from the fact that they have been proved very efficient in terms of sample complexity. Under the Cavender-Farris model of evolution (Cavender, 1978; Farris, 1973), Atteson (1999) showed that it is possible to guarantee that $\max_{i,j} |d_{ij} - \hat{d}_{ij}| < \epsilon$ with a probability at least $\delta$ if:

$$N \geq \frac{2 \ln\left(\frac{2n}{\delta}\right)}{(1 - \exp(-\epsilon))^2} \left( \exp\left( \max_{i,j} d_{ij} \right) \right)^2, \tag{24}$$

with $N$ the number of mutation sites (*i.e.* the number of observations) and $n$ the number of taxa (*i.e.* the number of OVs). Erdös et al. (1999) then demonstrated that under any evolutionary model and for any reconstruction method, $N$ grows at least as fast as $\log n$, and for any model assuming i.i.d. observations, it grows at least as $n \log n$. Erdös et al. also proposed a new algorithm, called Dyadic Closure Method, with a sample complexity of a power of $\log n$, when the mutation probabilities lie in a fixed interval. Daskalakis et al. (2009) proved the Steel's conjecture (Steel, 2001) which states that if the mutation probabilities on all edges of the tree are less than $p^* = (\sqrt{2} - 1)/2^{3/2}$ and are discretized, then the tree can be recovered in $\log n$. Recently, Mossel et al. (2011) proved that $\log^2 n$ suffices when discretization is not assumed.

In phylogenetics, the scientist is often faced with a set of different trees[11] and the construction of a consensus tree is thus required. The computational complexity of this construction has been studied and a polynomial algorithm has been proposed by Steel et al. (1992).

**Learning Dedicated to General LTM**

In this subsection, we present the latest developments of Choi et al. (2011) for general LTM learning, *i.e.* learning not restricted to phylogenetic trees. It is restricted to the analysis of data whose all variables share the same state space, for instance binary data. Assuming data generated by a parsimonious LTM, the additive tree metric property allows to exactly recover child-parent and sibling relations from distances (see Subsection Distances between Variables, last paragraph). Another advantage is that OVs are not necessarily constrained to be leaf nodes (this will be seen in the next paragraph).

The distance-based general LTM learning (**DBG**) is implemented in Algorithm 4, detailed as follows. First, the working node set **W** is initialized with the set of observed variables $\mathbf{X} = \{X_1, ..., X_n\}$ (line 1). Distances are computed for any three variables in **W** (line 2). An empty tree is created on **W** (line 3). Then the following steps are successively iterated (lines 4 to 16):

- A procedure (based on properties described in Subsection Distances between Variables, last paragraph) allows to identify nodes that have a parent-child relation and nodes that are siblings (line 5). This procedure generates three different sets of node groups: a set of parent-child groups, $\mathbf{PC} = \{PC_1, ..., PC_p\}$, a set of sibling groups, $\mathbf{S} = \{S_1, ..., S_q\}$ and a set of remaining single nodes, $\mathbf{R} = \{R_1, ..., R_r\}$;

- The content of the working node set **W** is replaced with parent nodes belonging to $parents(\mathbf{PC})$ and remaining single nodes belonging to **R** (line 6);

- For each group of sibling nodes $\in \mathbf{S}$, a new parent LV $H$ is created and added to **W** (lines 7 and 8);

- To update the distances of the working node set (line 9), the distances between the new LVs and the remaining variables in **W** are calculated (the calculation is easily derived from previously computed distances; for more details, see Choi et al., 2011);

---

11. For instance when different genes are used to infer trees.

**Algorithm 4** Distance-based general LTM learning (DBG, adapted from RG, Choi et al., 2011)

**INPUT:**
$\mathbf{X}$, a set of $n$ observed variables $\{X_1, ..., X_n\}$.

**OUTPUT:**
$T$ and $\theta$, respectively the tree structure and the parameters of the LTM constructed.

1: $\mathbf{W} \leftarrow \mathbf{X}$ /* Initialization of the working set of variables */
2: $\mathbf{D} \leftarrow info\_dist\_computation(\mathbf{W})$ /* for any three variables in $\mathbf{W}$ */
3: $T(\mathbf{V}, \mathbf{E}) \leftarrow empty\_tree(\mathbf{W})$ /* tree on $\mathbf{W}$ with no edges */
4: **loop**
5: $\quad (\mathbf{PC,S,R}) \leftarrow test\_relations(\mathbf{D,W})$ /* see paragraph 3 of the section */
6: $\quad \mathbf{W} \leftarrow (parents(\mathbf{PC}), \mathbf{R})$ /* parents and singles */
7: $\quad \mathbf{LCMT} \leftarrow LCM\_trees(\mathbf{S})$ /* an LCM tree for each sibling group */
8: $\quad \mathbf{W} \leftarrow \mathbf{W} \cup latent\_variables(\mathbf{LCMT})$
9: $\quad \mathbf{D} \leftarrow \mathbf{D} \cup info\_dist\_computation(\mathbf{W})$ /* only for LVs $\in \mathbf{W}$ */
10: $\quad \mathbf{E} \leftarrow \mathbf{E} \cup edges(\mathbf{PC}) \cup edges(\mathbf{LCMT})$
11: $\quad \mathbf{V} \leftarrow \mathbf{V} \cup latent\_variables(\mathbf{LCMT})$
12: $\quad$ **if** $|\mathbf{W}| = 2$ **then**
13: $\quad\quad \mathbf{E} \leftarrow \mathbf{E} \cup edge\_between\_the\_two\_remaining\_nodes(\mathbf{W})$
14: $\quad\quad$ **break**
15: $\quad$ **else if** $|\mathbf{W}| = 1$ **then break end if**
16: **end loop**
17: $\theta \leftarrow EM(T)$ /* See Section 3.1 */

- If the working node set $\mathbf{W}$ contains strictly more than two nodes, then a new step is started. Otherwise, there are two possible situations: if the working node set $\mathbf{W}$ contains two nodes, then these nodes are connected[12] and the procedure is completed (lines 12 to 14); if the working node set $\mathbf{W}$ only contains one node, then the iteration stops (line 15).

After learning the partial structure, model parameters (line 17) are learned (see Section 3.1).

In practice, Choi et al. (2011) restricted the learning to symmetric discrete distributions[13] (see definition in Subsection Distances between Variables). This restriction presents a major advantage: it allows to derive model parameters through the use of distances previously learned by structure recovery. In other words, after learning the structure, there is no need to recover parameters through EM. This is due to the one-to-one correspondence between distance and model parameters (for more details, see Choi et al., 2011).

---

12. It prevents the introduction of a redundant LV, see Section 2.5, second paragraph.
13. Nevertheless, the algorithm can be applied to non-symmetric discrete distributions. The only requirement is that all variables share the same state space.

To diminish the computational complexity of DBG, Choi et al. (2011) propose to first learn a minimum spanning tree (MST) based on distances between OVs. Then, in the tree, they identify the set of internal nodes. For each internal node $V_i$ and its neighbor nodes $nbd(V_i)$, they apply DBG which outputs a latent tree. In the global model, each subtree $\{V_i, nbd(V_i)\}$ is then replaced by the corresponding latent tree. This strategy allows to reduce the computational complexity of latent tree construction because MST is fast to compute and DBG is only applied to a restricted number of variables. In term of sample complexity, DBG and its derived algorithms only require $\log n$ observations for recovering the model with high probability. This sample complexity is equal to the one of the Chow-Liu algorithm (Tan et al., 2011).

Another version of the DBG algorithm was developed when it is not assumed that observations have been generated by a genuine LTM. To prevent the incorporation of irrelevant LVs, after applying DBG on subtrees $\{V_i, nbd(V_i)\}$, only are integrated in the model those latent trees that increase the BIC score.

**Spectral Methods**

Recent works extended previous distance methods following a spectral approach. On the one hand, Anandkumar et al. (2011) addressed the multivariate setting where observed and latent nodes are random vectors rather than scalars. Their approach can deal with general linear models containing both categorical and continuous variables. Another important improvement of their method is that sample complexity bound is given in terms of natural correlation conditions that generalize the more restrictive effective depth conditions of previous works (Erdos et al., 1999; Choi et al., 2011). The proposed extension consists in replacing the step in line 5 of Algorithm 4 by a quartet test[14] relying on spectral techniques (more specifically, canonical correlation analysis; Hair, Black, Babin, and Anderson, 2009). Given four observed variables $\{X_1, X_2, X_3, X_4\}$, the spectral quartet test distinguishes between the four possible tree topologies (see Figure 8). The correct topology is $\{\{X_i, X_j\}, \{X_{i'}, X_{j'}\}\}$ if and only if:

$$|E[X_i X_j^\top]|_* \ |E[X_{i'} X_{j'}^\top]|_* > |E[X_{i'} X_j^\top]|_* \ |E[X_i X_{j'}^\top]|_*, \tag{25}$$

where $|M|_* := \Pi_{\ell=1}^k \sigma_\ell(M)$ is the product of the k largest singular values of matrix $M$ and $E[M]$ is the expectation of $M$ (estimated using the covariance matrix).

On the other hand, Song et al. (2011) proposed a non-parametric learning based on kernel density estimation (KDE) (Rosenblatt, 1956; Parzen, 1962). KDE is particularly relevant for model learning, in the case of non-Gaussian continuous variables showing multimodality and skewness. Given a set of $N$ i.i.d. observed data $D_{\mathbf{x}} = \{\mathbf{x}^1, ..., \mathbf{x}^N\}$, the joint distribution is modeled as :

$$P(\mathbf{x}) = \frac{1}{N} \sum_{i=1}^N \prod_{j=1}^n k(x_j, x_j^i), \tag{26}$$

with $\mathbf{x} = \{x_1, ..., x_n\}$ an observation, $k(x, x')$ the Gaussian radial basis function and $N$ the number of observations. Kernels $k(x, x')$ are represented as inner products $\langle \phi(x), \phi(x') \rangle_{\mathcal{F}}$ through a feature map $\phi : \mathbb{R} \to \mathcal{F}$. As products of kernels are also kernels, the product $\Pi_{j=1}^n k(x_j, x'_j)$ is written as a single inner product $\langle \otimes_{j=1}^n \phi(x_j), \otimes_{j=1}^n \phi(x'_j) \rangle_{\mathcal{F}^n}$, where $\otimes$ is the tensor product. Let $\mathcal{C}_\mathbf{X} := E_\mathbf{X}[\otimes_{j=1}^n \phi(X_j)]$ be the Hilbert space embedding of the KDE distribution $P(\mathbf{X})$. The expectation of $P(\mathbf{X})$ can be formulated as $\langle \mathcal{C}_\mathbf{X}, \otimes_{j=1}^n \phi(x_j) \rangle_{\mathcal{F}^n}$. By exploiting a given latent tree structure, the key is that Hilbert space embedding allows to decompose $\mathcal{C}_\mathbf{X}$ into simpler tensors. Similarly to the work of Parikh et al. (2011), message passing and parameter estimation are reformulated through tensor notation (see the work of Song et al., 2011, for more details). To learn the structure, a non-parametric additive tree metric is used. For two variables $V_i$ and $V_j$, the distance is:

$$d_{ij} = -\frac{1}{2} \log |\mathcal{C}_{ij} \mathcal{C}_{ij}^\top|_\star + \frac{1}{4} \log |\mathcal{C}_{ii} \mathcal{C}_{ii}^\top|_\star + \frac{1}{4} \log |\mathcal{C}_{jj} \mathcal{C}_{jj}^\top|_\star. \qquad (27)$$

---

14. Quartet tests are widely used for phylogenetic tree inference. The first authors to adapt them to LTM learning were Chen and Zhang (2006).

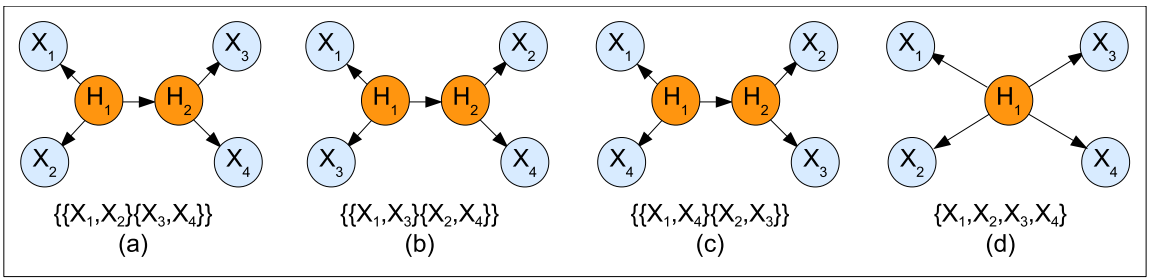

Figure 8: The four possible tree topologies for the quartet test. See Figure 2 for node color code.

Then, given the distances between variables, NJ or DBG can be used to learn the structure.

### 3.2.4 Determination of Latent Variable Cardinalities

During LCM learning, LV cardinality can be determined through the examination of all possible values (up to a maximum) and then by choosing the one which maximizes a criterion, such as BIC (Raftery, 1986). For each cardinality value, parameters are required to calculate the likelihood appearing in the optimization criterion. For this purpose, random restarts of EM are generally used to learn parameters with a low probability of getting trapped in local maxima. The drawback is that this method cannot be applied to LTM learning, because EM becomes time-consuming when there are several LVs. A better solution consists in using a greedy search approach, starting with a preset value of LV cardinality (generally equal to 2) and incrementing it to meet the optimal criterion. However, this solution still remains computationally demanding (Zhang, 2004).

To tackle the issue of computational burden, several strategies have been proposed. For instance, one can simply set a small value for the LV cardinalities. Following this idea, Hwang and collaborators (2006) constrain LVs to binary variables. Because they worked on binary trees whose OVs are also binary, this restriction is not severe in practice. Nevertheless, in the case of non-binary OVs and/or non-binary trees, this very fast method presents several drawbacks: on the one hand, a too small cardinality can lead to an important loss of information; on the other hand, a too large cardinality can entail model overfitting and unnecessary computational burden.

More rigorous, Wang et al.'s approach (2008) relies on regularity (see Section 2.5). Knowing the cardinality of its neighbor variables $Z_i$, the cardinality of an LV $H$ is determined as follows:

$$|H| = \frac{\Pi_{i=1}^{k} |Z_i|}{\max_{i=1}^{k} |Z_i|}. \tag{28}$$

This very fast approach is efficient for LVs owning a few number of neighbors. Thus this is only practicable for binary trees or trees whose LV degrees are small (close to 3). In the context of large scale data analysis (several thousands of variables), Mourad et al. (2011) proposed to estimate the cardinality of an LV given the number of its children. The rationale underlying this approach is the following: the more child nodes an LV has, the larger the number of combinations is for the values of the child variables. Therefore, the cardinality of a latent variable should depend on the number of its child nodes. Nonetheless, to keep the model complexity within reasonable limits, a maximum cardinality is fixed.

Two additional methods have been proposed to offer a better trade-off between accuracy and computational cost. The first one uses a search-based agglomerative state-clustering procedure (Elidan & Friedman, 2001). The idea relies on the Markov blanket of an LV. In LTMs, the Markov blanket of an LV $H$, noted $\mathbf{MB}_H$, is composed of its parent and its children. The Markov blanket represents the set of variables that directly interact with $H$. Elidan and Friedman's method sets the initial cardinality of $H$ based on the empirical joint distribution of $\mathbf{MB}_H$, noted $\hat{P}(\mathbf{MB}_H)$. $H$ is initialized to have a state for each configuration found in $\hat{P}(\mathbf{MB}_H)$. Then the cardinality is repeatedly decreased through successive merging operations: states $h_i$ and $h_j$, whose merging entails the best optimization of a given score, are merged. After repeating these operations till $H$ has only one state, the cardinality value leading to the best score is selected. The second method relies on local and fast parameter estimation through LCM learning (Harmeling & Williams, 2011). As presented in the first paragraph of this section, a greedy search approach can be used. It starts with a preset value and increments it to meet an optimal criterion. This greedy search becomes efficient because, for each cardinality value to test, parameters are quickly learned in constant time.

### 3.2.5 Choosing a Root

The LTM root cannot be learned from data. However there is sometimes a need to determine the root. For instance, LCM-based parameter learning (see Algorithm 1) can easily be performed when a root is chosen.

The root can be determined from *a priori* knowledge on data. For instance, we can consider that the latent structure of LTM represents a hierarchy of concepts (*i.e.* a taxonomy in ontology). Thus, the LV root corresponds to the highest abstract level, whereas an LV node only having OVs as children is interpreted as the lowest abstract level. Actually, variable clustering-based algorithms implicitly implement this *a priori* knowledge.

### 3.2.6 Time Complexity and Scalability

The time complexity of generic LTM learning algorithms is summarized in Table 1. In the table, we compare algorithms, approaches, models and time complexities. We also give examples of instantiations for generic algorithms. Online resources are summarized in Appendix A. In order to simplify the comparison of time complexities, we only consider the number $n$ of variables (input data), the number $N$ of observations and the number $s$ of steps (for search-based algorithms). The LTM learning algorithms are compared with the Chow-Liu algorithm for learning a tree without LVs. Details about the complexity calculation of LTM learning algorithms are provided in Supplemental material B.2.

| Algorithm | Approach | Model | Complexity | Instantiation |
|---|---|---|---|---|
| CL | - | Tree | $O(n^2N)$ | – (Chow & Liu, 1968) |
| NGS | Score | Tree | $O(sn^5N)$ | DHC (Zhang, 2004) |
| AGS, Alg. 2 | Score | Tree | $O(sn^2N)$ | HSHC (Zhang & Kocka, 2004b)<br>EAST (Chen et al., 2012) |
| AHCB | Variable clustering | Forest | $O(n^2N)$ | LTAB (Wang et al., 2008)<br>BIN-A (Harmeling & Williams, 2011) |
| LCMB-LTM, Alg. 3 | Variable clustering | Forest | $O(n^2N)$ | BIN-G (Harmeling & Williams, 2011)<br>CFHLC (Mourad et al., 2011)<br>BI (Liu et al., 2012) |
| NJ | Information distance | Tree | $O(n^3N)$ | NINJA (Saitou & Nei, 1987; Wheeler, 2009) |
| DBG, Alg. 4 | Information distance | Tree | $O(n^3N + n^4)$<br>$O(n^2N + n^4)$ | RG (Choi et al., 2011)<br>CLRG (Choi et al., 2011)<br>regCLRG (Choi et al., 2011) |

Table 1: Computational time complexities of generic algorithms dedicated to latent tree model (LTM) learning. The number of observed variables, the number of observations and the number of steps (in search-based algorithms) are denoted $n$, $N$ and $s$, respectively. CL: Chow-Liu's algorithm; NGS: naive greedy search (Section 3.2.1); AGS: advanced greedy search (Section 3.2.1); LCMB-LTM: latent class model-based LTM learning (Section 3.2.2); NJ: neighbor joining (Section 3.2.3); DBG: distance-based general LTM learning (Section 3.2.3).

When the tree does not contain any LV, learning the model can be done efficiently in $O(n^2N)$ using the Prim's algorithm (1957). The situation is more complicated when the tree contains LVs. The complexity of finding the regular LTM with the lowest score is $O(2^{3n^2})$. Search-based methods implement heuristics to reduce this large complexity. Their overall complexity can be decomposed into a product of three main terms: number of steps, structure learning complexity and parameter learning complexity. At the opposite, in variable clustering- and distance-based methods, the overall complexity can be decomposed into a sum. Nevertheless, the development of new operators for greedy search and the application of local EM have led to significant improvements (from $O(sn^5N)$ to $O(sn^2N)$). Variable clustering-based methods are computationally more efficient for multiple reasons. They rely on pairwise dependence computation to identify LVs and their connections, and on LCM-based learning to determine LV cardinality. Regarding distance-based methods, NJ provides a reasonable complexity of $O(n^3N)$, whereas DBG presents a high complexity of $O(n^3N+n^4)$. However, this last complexity corresponds to the worst case, when the tree to learn is a hidden Markov model. Besides, Choi et al. provide a modified DBG which reduces the complexity to $O(n^2N + n^4)$.

| Algorithm | BinTree | BinForest | Asia | Hannover | Car | Tree |
|---|---|---|---|---|---|---|
| Number of variables | 4 | 5 | 8 | 5 | 7 | 19 |
| Number of observations | 500 | 500 | 100 | 3589 | 869 | 500 |
| Type of data | simu | simu | simu | real | real | simu |
| CL[1] | 0.01 | 0.01 | 0.01 | 0.00 | 0.01 | 0.04 |
| LCM | 1.43 | 5.83 | 1.02 | 58.58 | 2.54 | 1.61 |
| DHC | 18.79 | 123.83 | 16.23 | 9.86 | 1609.72 | *time* |
| SHC | 27.07 | 43.86 | 4.55 | 5.39 | 150.23 | 4258.06 |
| HSHC | 13.85 | 15.14 | 1.5 | 1.9 | 18.79 | 87.04 |
| EAST | 12.56 | 20.17 | 3.79 | 5.02 | 63.55 | 309.47 |
| LTAB | 2.8 | 5.97 | 0.97 | 1 | 35.36 | 86.52 |
| BIN-A | 3.00 | 2.68 | 0.22 | 16.70 | 3.72 | 0.63 |
| BIN-G | 3.06 | 2.61 | 0.23 | 17.87 | 3.72 | 0.29 |
| CFHLC | 1 | 1.3 | 0.6 | 18.6 | 2.7 | 6.6 |
| BI | 19.38 | 37.66 | 7.87 | 7.65 | 69.26 | 183.34 |
| NJ | *non-bin* | *non-bin* | 0.15 | 4.75 | *non-bin* | *non-bin* |
| RG[2] | *non-bin* | *non-bin* | 0.16 | 3.49 | *non-bin* | *non-bin* |
| CLRG[2] | *non-bin* | *non-bin* | 0.06 | 3.63 | *non-bin* | *non-bin* |
| regCLRG[2] | *non-bin* | *non-bin* | 0.04 | 6.54 | *non-bin* | *non-bin* |
| Algorithm | Forest | Alarm | Coil-42 | NewsGroup | HapGen | HapMap |
| Number of variables | 20 | 37 | 42 | 100 | 1000 | 10000 |
| Number of observations | 500 | 1000 | 4000 | 8121 | 1000 | 116 |
| Type of data | simu | simu | real | real | simu | real |
| CL[1] | 0.04 | 0.15 | 0.45 | 2.17 | 121.36 | *memory* |
| LCM | 0.97 | 34.32 | 678.19 | 1467.10 | *bug* | *memory* |
| DHC | *time* | *time* | *time* | *time* | *time* | *time* |
| SHC | 3729.78 | *time* | *time* | *time* | *time* | *time* |
| HSHC | 158.64 | 4366.93 | *time* | *time* | *time* | *time* |
| EAST | 277.01 | 6388.69 | 75168[3] | *time* | *time* | *time* |
| LTAB | 83.31 | 574.57 | 2197.69 | *time* | *time* | *time* |
| BIN-A | 2.03 | 86.12 | 387.21 | 1152.70 | 3573.20 | *memory* |
| BIN-G | 1.28 | 99.93 | 436.41 | 1302.10 | 7671.20 | *memory* |
| CFHLC | 5.4 | 21 | 560.9 | 1291.4 | 787.2 | 2852.6 |
| BI | 223.19 | 319.62 | 1193.25 | 6311.99 | 18977.02 | *time* |
| NJ | *non-bin* | *non-bin* | *non-bin* | 1325.38 | *non-bin* | *memory* |
| RG[2] | *non-bin* | *non-bin* | *non-bin* | 274.37 | *non-bin* | *memory* |
| CLRG[2] | *non-bin* | *non-bin* | *non-bin* | 927.22 | *non-bin* | *memory* |
| regCLRG[2] | *non-bin* | *non-bin* | *non-bin* | 345.09 | *non-bin* | *memory* |

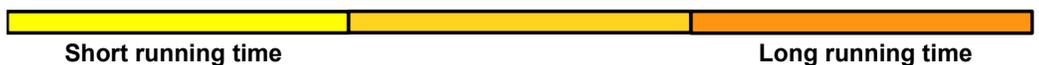

**Short running time**        **Long running time**

Table 2: Comparison of running times between algorithms from the literature on small, large and very large simulated and real datasets. 1: CL learns a tree without LVs; 2: RG, CLRG and regCLRG learn a tree whose internal nodes can be observed or latent. *time:* very long running time; *memory:* out-of-memory; *non-binary:* impossible to process non-binary data. 3: results from the work of Chen (2008).

183

| Algorithm | BinTree | BinForest | Asia | Hannover | Car | Tree |
|---|---|---|---|---|---|---|
| Number of variables | 4 | 5 | 8 | 5 | 7 | 19 |
| Number of observations | 500 | 500 | 100 | 3589 | 869 | 500 |
| Type of data | simu | simu | simu | real | real | simu |
| CL[1] | -3222.8±0 | -3350.2±0 | -281.43±0 | -7859.6±0 | -7161.2±0 | -10628±0 |
| LCM | -3361.8±30 | -3646.9±8 | -346.23±10 | -7754.6±3 | -7127.8±18 | **-10010**±0 |
| DHC | **-2825.11**±0 | **-3038.66**±0 | -269.32±6 | -7710.96±1 | **-7049.98**±22 | time |
| SHC | **-2825.11**±0 | -3056.77±0 | -268.04±4 | -7709.71±0 | -7056.18±4 | -10095.92±17 |
| HSHC | -2825.1±0 | -3056.77±0 | -270.59±0 | -7709.71±0 | -7057.49±3 | -10087.54±8 |
| EAST | -2825.11±0 | -3056.76±0 | -283.82±6 | **-7709.69**±0 | -7051.97±7 | -10092.45±5 |
| LTAB | -3332.94±8 | -3791±47 | -727.3±23 | -7876.48±0 | -8084.88±57 | -13410.27±390 |
| BIN-A | -2991±0 | -3146±0 | -296.01±0 | -7756±0 | -7137.7±41 | **-10010**±0 |
| BIN-G | -2991±0 | -3146±0 | -296.01±0 | -7756±0 | -7133.8±43 | **-10010**±0 |
| CFHLC | -3682.86±0 | -3302.56±0 | -280.86±2 | -8032.39±0 | -7200.76±26 | -10070.43±10 |
| BI | -2850±0 | -3074±0 | -283±0 | -7711±1 | -7063±17 | -10073±2 |
| NJ | non-bin | non-bin | -288.39±1 | -7714±3 | non-bin | non-bin |
| RG[2] | non-bin | non-bin | -287.13±2 | -7711.4±4 | non-bin | non-bin |
| CLRG[2] | non-bin | non-bin | -271.81±0 | -7710.1±2 | non-bin | non-bin |
| regCLRG[2] | non-bin | non-bin | **-266.14**±0 | -7742.4±23 | non-bin | non-bin |
| **Algorithm** | **Forest** | **Alarm** | **Coil-42** | **NewsGroup** | **HapGen** | **HapMap** |
| Number of variables | 20 | 37 | 42 | 100 | 1000 | 10000 |
| Number of observations | 500 | 1000 | 4000 | 8121 | 1000 | 116 |
| Type of data | simu | simu | real | real | simu | real |
| CL[1] | -11330±0 | **-11281**±0 | **-36444**±0 | -121400±0 | **-191250**±0 | memory |
| LCM | -10709±0 | -21859±3489 | -49581±491 | -124390±768 | bug | memory |
| DHC | time | time | time | time | time | time |
| SHC | -10777.81±1 | time | time | time | time | time |
| HSHC | -10775.25±0 | -11322.83±84 | time | time | time | time |
| EAST | -10777.08±5 | -12315.66±589 | -35982.4[3] | time | time | time |
| LTAB | -14207.87±392 | -17733.87±239 | -43655.37±46 | time | bug | time |
| BIN-A | **-10708**±0 | -17640±551 | -37380±104 | -119060±33 | -301010±3962 | memory |
| BIN-G | **-10708**±0 | -17600±589 | -37404±78 | -120230±231 | -301790±3006 | memory |
| CFHLC | -10762.3±8 | -17856.04±18 | -51878.73±151 | -129101.4±488 | -367875.3±1706 | **-373523**±876 |
| BI | -10761±6 | -12296±125 | -36682±152 | **-117278**±134 | -267881±1347 | time |
| NJ | non-bin | non-bin | non-bin | -117610±39 | non-bin | memory |
| RG[2] | non-bin | non-bin | non-bin | -120274±380 | non-bin | memory |
| CLRG[2] | non-bin | non-bin | non-bin | -117580±46 | non-bin | memory |
| regCLRG[2] | non-bin | non-bin | non-bin | -118938±161 | non-bin | memory |

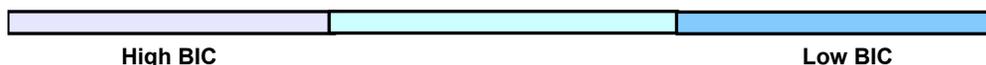

High BIC     Low BIC

Table 3: Comparison of BIC scores between algorithms from the literature on small, large and very large simulated and real datasets. 1: CL learns a tree without LVs; 2: RG, CLRG and regCLRG learn a tree whose internal nodes can be observed or latent. *time:* very long running time; *memory:* out-of-memory; *non-binary:* impossible to process non-binary data. 3: results from the work of Chen (2008).

Some of the algorithms proposed in the literature differ from the generic algorithms presented. In Tables 2 and 3, we compare the algorithms from the literature on small, large and very large simulated and real datasets[15]. Moreover we provide results for standard algorithms: CL model-based and LCM-based approaches. For each dataset, we learned the model from training data and evaluated the BIC score on test data. We repeated the experiments 10 times. The programs were allowed to run in maximum 6 hours. Datasets are described in Supplemental material B.3[16]. We report the BIC score with its standard deviation and the running time.

On small datasets ($n \leq 10$ variables), search-based methods lead to the best BIC values, except for the Asia dataset for which a distance-based method, regCLRG, is the best one. This is not surprising since search-based methods evaluate a large number of models to find the optimal one. Nevertheless, on large datasets ($10 \leq n \leq 100$), search-based methods require long running times and thus cannot be used for some datasets such as the Coil-42 and NewsGroup ones. In this context, variable clustering-based and distance-based methods are much more efficient while yielding accurate results. Regarding the very large dataset context ($n > 100$), only variable clustering-based and distance-based methods can learn LTMs[17]. CFHLC[18] is the only approach able to process the HapMap data containing 117 observations for $10k$ variables. For all datasets, we observe that using CL model and LCM predominantly leads to lower BIC scores than when using LTM, except for large and very large datasets.

### 3.2.7 Summary

LTM learning has been subject to many methodological developments. When structure is known, EM is often preferred. Nevertheless, for large LTMs, EM leads to considerable running times and to local maxima. To address this problem, LCM-based EM allows to quickly learn parameters, while spectral methods help find the optimal solution when LTM parameters are not required. When structure is unknown, search-based approaches represent standard methods from Bayesian network learning. However they are only suitable for learning small LTMs. To tackle this issue, variable clustering-based methods represent efficient alternatives. These methods are based on the idea of grouping variables to identify LVs in a bottom-up manner. Recently, phylogenetic algorithms have been adapted to general LTM learning. Compared to the other methods, they guarantee to exactly recover the generative LTM structure under some conditions.

---

15. For a fair comparison, we used the implementation of NJ provided by Choi et al. (2011).
16. Although algorithms NJ, RG, CLRG and regCLRG can process any kind of data with shared state space (binary data, ternary data, ...), the implementation provided by Choi et al. (2011) can only process binary data. Hence the algorithms have not been applied to some datasets. We recall that RG, CLRG and regCLRG do not exactly learn an LTM but instead a tree whose all internal nodes are not compelled to be latent.
17. Although it is not shown in Tables 2 and 3, NJ, RG, CLRG and regCLRG were able to process 1000 *binary* variables in our experiments.
18. In the work of Mourad et al. (2011), CFHLC implements a window-based approach to scale very large datasets ($n \geq 100k$ variables). Here for a fair comparison, the window-based approach has not been used.

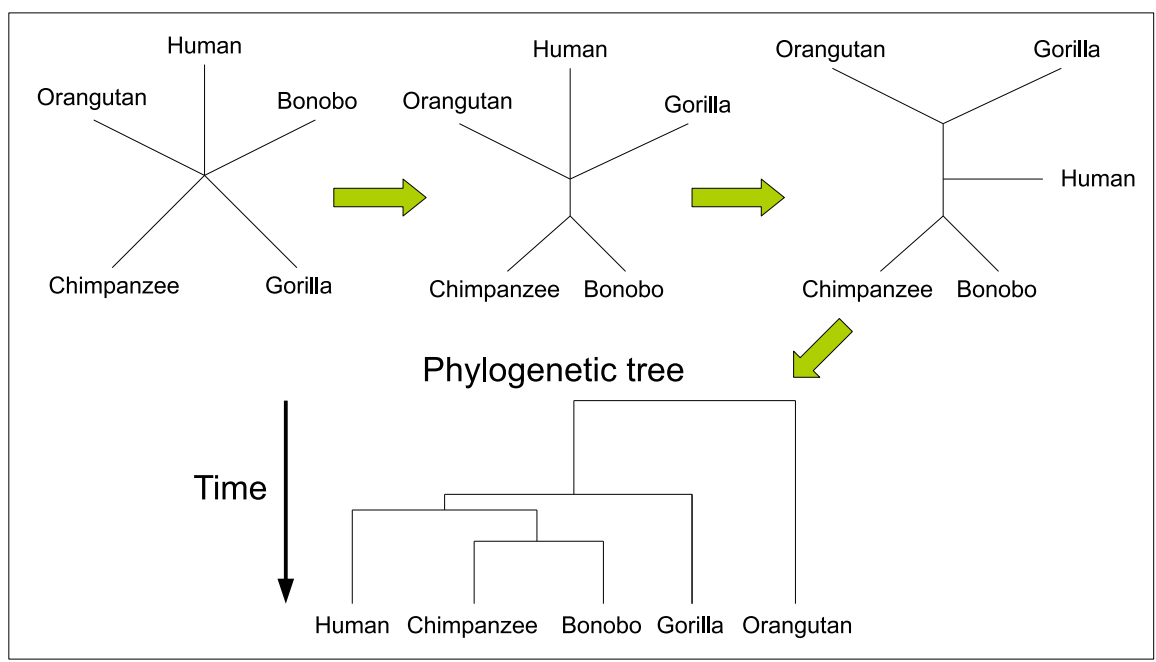

Figure 9: Illustration of phylogenetic tree reconstruction.

## 4. Applications

In this section, we discuss and illustrate three types of applications of LTMs: latent structure discovery, multidimensional clustering and probabilistic inference. At the end of the section, we also briefly present other applications such as classification.

### 4.1 Latent Structure Discovery

Latent structure discovery aims at revealing: (i) latent information underlying data, *i.e.* unobservable variables or abstract concepts which have a role to play in data analysis, and (ii) latent relationships, *i.e.* relationships existing between observed and latent information, and also between pieces of latent information themselves. For this purpose, LTM analysis represents a powerful tool where latent information and latent relationships are modeled by LVs and graph edges, respectively. Thanks to LTMs, latent structure discovery has been applied to several fields: marketing (Zhang, Wang, & Chen, 2008), medicine (Zhang, Nielsen, & Jensen, 2004; Zhang et al., 2008), genetics (Hwang et al., 2006; Mourad et al., 2011) and phylogenetics (Felsenstein, 2003; Friedman, Ninio, Pe'er, & Pupko, 2002). Let us take the example of phylogenetics which is the major application of LTMs in structure discovery.

In phylogenetics, the purpose is to infer the tree representing the evolutionary connections between observed species. Let us consider human and its closest living relatives: orangutan, gorilla, bonobo and chimpanzee. From their DNA sequences, it is possible to reconstruct the phylogenetic tree. DNA sequences are sequences of letters $A$, $T$, $G$ and $C$. During the evolution of species, DNA sequences are modified by mutational processes. Each species can then be characterized by its DNA sequence. One of the most popular algorithms for phylogenetic tree reconstruction is NJ (described in Section 3.2.3, Neighbor Joining). It starts by considering the tree as a star linking all species (see illustration in Figure 9). Then the distances between all species are calculated based on the DNA sequences. Chimpanzee and bonobo present the shortest distance and are thus regrouped under a new latent node. Then distances are updated and the last previous step is reiterated until the construction of the final phylogenetic tree. The tree first links chimpanzee and bonobo, then human, gorilla and orangutan. The success of NJ comes from the fact that, compared to previous hierarchical clustering methods such as UPGMA (Unweighted Pair Group Method with Arithmetic Mean), it does not assume all species evolve at the same rate. The length of an edge represents the time separating two species. Moreover, assuming that the distances are correct, NJ outputs the correct tree.

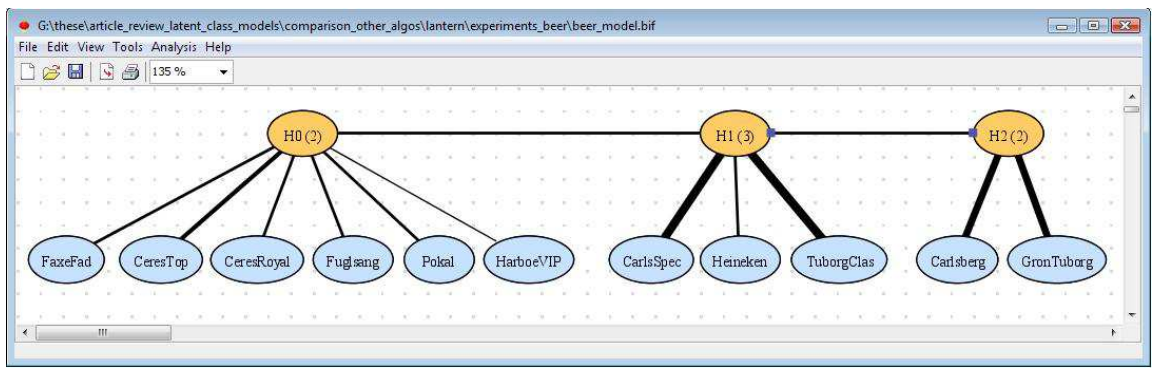

Figure 10: Latent tree model learned from the dataset on Danish beer consumption. Edge widths are proportional to mutual information between nodes. For each latent variable, the number of latent classes is indicated in brackets. See Figure 2 for node color code. Lantern software.

### 4.2 Multidimensional Clustering

Cluster analysis, also called clustering, aims at assigning a set of observations to several groups (called clusters) so that observations belonging to the same cluster are similar in some sense (Xu & Wunsch, 2008). LTMs are particular tools which can produce multiple clusterings: each LV represents a partition of data, which is most related to a specific subset of variables. This application is called "multidimensional clustering" and has been mainly explored by Chen et al. (2012).

Let us illustrate LTM-based multidimensional clustering using dataset from a survey on Danish beer market consumption. For this purpose, we use the user-friendly software Lantern. The dataset consists of 11 variables and 463 consumers. Each variable represents a beer brand which is evaluated through the four possible answers to a survey questionnaire: never seen the brand before ($s0$); seen before, but never tasted ($s1$); tasted, but do not drink regularly ($s2$) and drink regularly ($s3$).

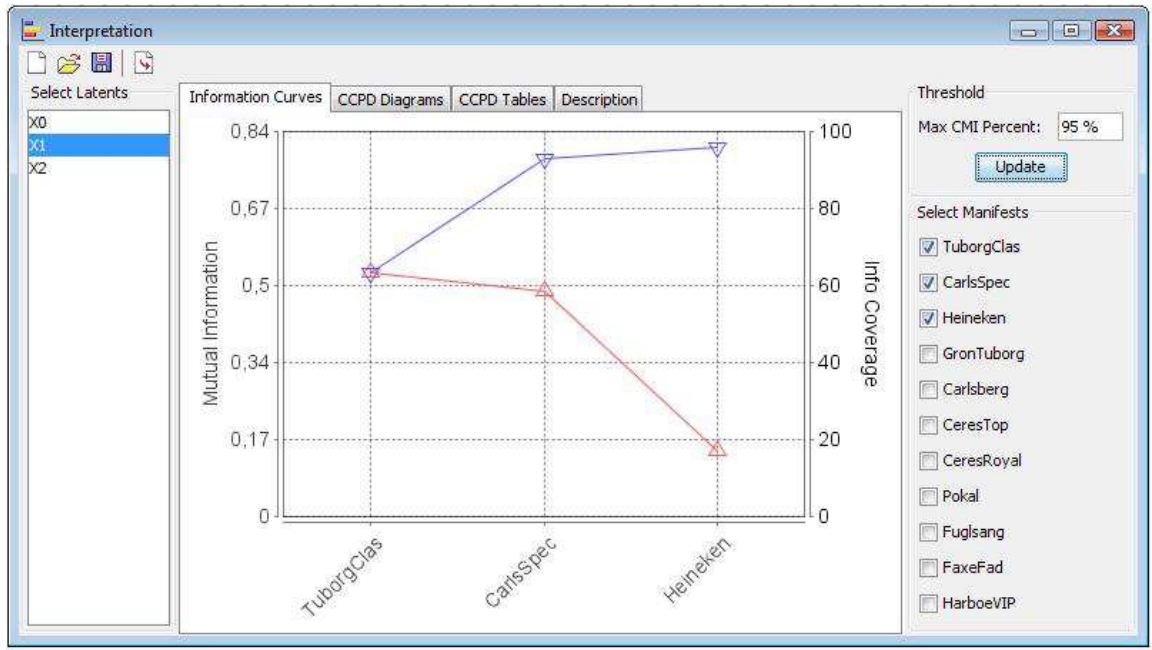

Figure 11: Information curves for latent variable $H1$. Lantern software.

The learned model is presented in Figure 10. It has a BIC score of $-4851.99$. The model contains 3 LVs: $H_0$, $H_1$ and $H_2$. These LVs have 2, 3 and 2 latent classes, respectively. Let us start with $H_1$ which is the simplest interpretable LV. Let $X_1, X_2, ..., X_n$ be the OVs sorted by decreasing values of pairwise MI between $H_1$ and each OV $X_i$. Two different information curves are analyzed in Figure 11: (i) the curve of pairwise mutual information $I(H_1; X_i)$ between $H_1$ and each OV $X_i$, and (ii) the curve of cumulative information $I(H_1; X_1, ..., X_i)$ representing MI between $H_1$ and the first $i$ OVs $X_1, ..., X_i$. The curve of pairwise mutual information shows that TuborgClas, followed by CarlSpec and Heineken, are the beers most related to $H_1$. The curve of cumulative information presents complementary information. The cumulative information curve increases monotonically with $i$ and reaches the maximum for $n$. The ratio $I(H_1; X_1, ..., X_i)/I(H_1; X_1, ..., X_n)$ is the information coverage of the first $i$ OVs. If this ratio is equal to 1, it means that $H_1$ is conditionally independent of $X_{i+1}, ..., X_n$ given the first $i$ OVs. In practice only the first OVs whose information coverage is less than 95% are considered relevant. Using the cumulative information curve, we observe that $H_1$ is only related to TuborgClas, CarlSpec and Heineken, which represent a group of beers different from the others. TuborgClas and CarlSpec are frequent beers, being a bit darker in color and more different in taste than the two main mass-market beers, GronTuborg and Carlsberg. Although not Danish, Heineken is one of the largest brand in the world that most Danes would have tasted sometimes during travels abroad. Results for other LVs are discussed but not shown (interpretation remains the same as for $H_1$). $H_0$ is more related to CeresTop, CeresRoyal, Pokal, Fuglsang, CarlsSpec and FaxeFad (*i.e.* minor local beers), whereas $H_2$ is more connected to GronTuborg and Carlsberg (*i.e.*, the two main mass-market beers).

|     | **Class1**, prior = 0.36 | | | **Class2**, prior = 0.27 | | | **Class3**, prior = 0.37 | | |
| --- | --- | --- | --- | --- | --- | --- | --- | --- | --- |
|     | Tub | Carl | Hein | Tub | Carl | Hein | Tub | Carl | Hein |
| $s0$ | 0.03 | 0 | 0.08 | 0.06 | 0.15 | 0.36 | 0 | 0 | 0.02 |
| $s1$ | 0.07 | 0.12 | 0.3 | **0.56** | **0.74** | **0.4** | 0 | 0.01 | 0.17 |
| $s2$ | **0.89** | **0.81** | **0.57** | 0.32 | 0.11 | 0.23 | 0.14 | 0.39 | **0.66** |
| $s3$ | 0.02 | 0.07 | 0.05 | 0.06 | 0 | 0.01 | **0.86** | **0.61** | 0.16 |

Table 4: Class conditional probability tables for latent variable $H1$. Tub: TuborgClas; Carl: CarlsSpec; Hein: Heineken.

|     |     | $H_1$ | | |
| --- | --- | --- | --- | --- |
|     |     | **Class1** | **Class2** | **Class3** |
| $H_2$ | **Class1** | **0.55** | **0.71** | 0.11 |
|     | **Class2** | 0.45 | 0.29 | **0.89** |

Table 5: Conditional probability distributions of latent variable $H_2$ given $H_1$.

Class conditional probability distributions (CCPDs) of $H_1$ are presented in Table 4. Using this table, it is easy to interpret latent classes. For instance, the first class (**Class1**) represents 36% of the consumers (prior = 0.36). For this class, all conditional probabilities of $s2$ are higher than 0.5. This means that these consumers tasted the beers, but do not drink them regularly. The second class (**Class2**) contains 27% of the consumers and represents people who saw the beers before or only tasted them. The last class (**Class3**) includes 37% of the consumers and represents people who just tasted the beers or drink them regularly. Results for other LVs are discussed but not shown. The CCPDs relative to $H_0$ show a division into a group of consumers who just tasted the beers (**Class1**) and a more complicated group of consumers who never saw the brands, just saw them or just tasted them (**Class2**). Regarding $H_2$, the CCPDs report consumers who just tasted the beers (**Class1**) and consumers who drink them regularly (**Class2**). Using the LTM, we can also analyze relations between the different partitions. The conditional probability distribution $P(H_2|H_1)$ is given in Table 5. We observe that consumer behaviors are similar for the two groups of beers. For instance, consumers who just tasted or regularly drink the $H_1$ group of beers (**Class3** of $H_1$) generally also drink regularly the $H_2$ group of beers (**Class2** of $H_2$).

In this example, we were able to find consumer profiles specific to beer brands. Multidimensional clustering thanks to LTMs helps discover multiple facets of data (*i.e.* LVs) and partition data along each facet (*i.e.* latent class). Moreover, general relations between multiple facets are highlighted through connections between LVs.

### 4.3 Probabilistic Inference

Probabilistic inference is the process of answering probabilistic queries of the form $p(x|y)$, for an event $x$ given some knowledge $y$ (Koller & Friedman, 2009), using the Bayes formula:

$$p(x|y) = \frac{p(y|x)p(x)}{p(y)}. \tag{29}$$

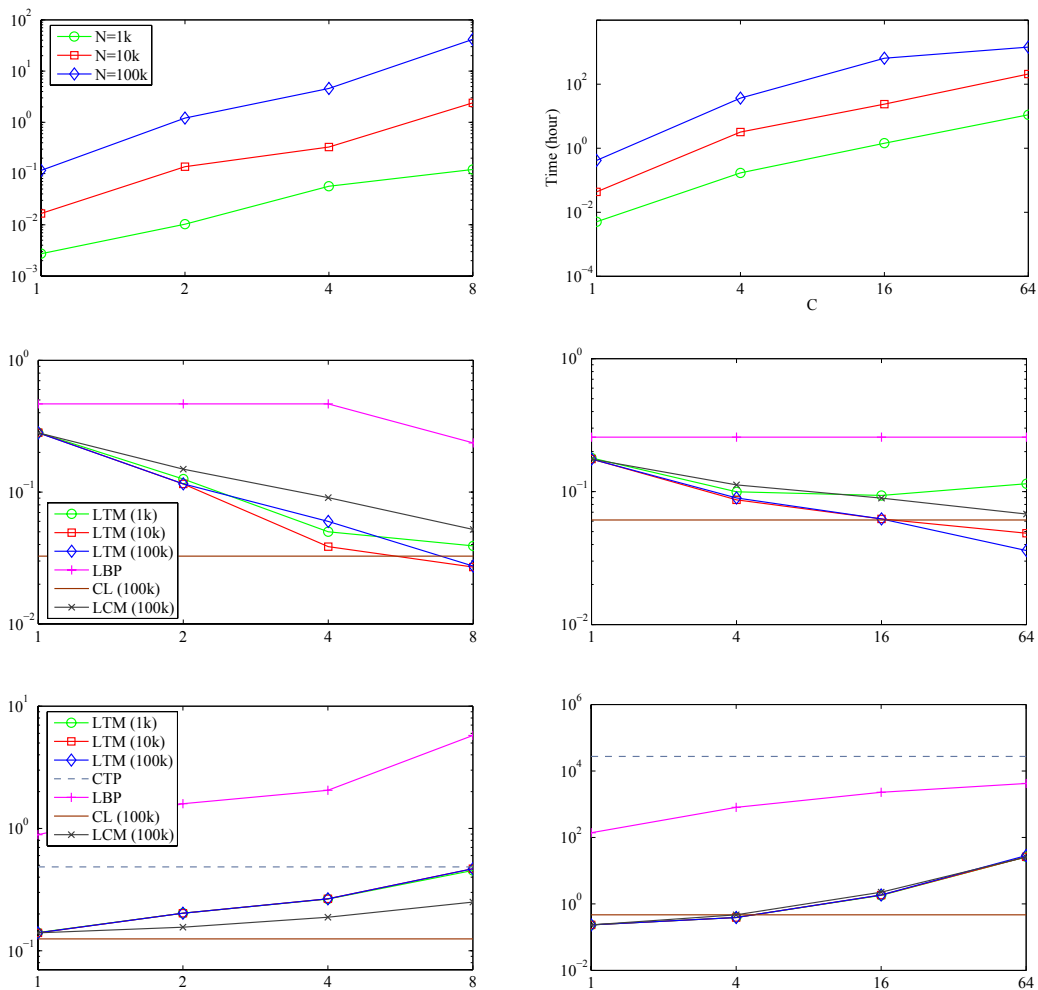

Figure 12: Experiments on ALARM and BARLEY networks: a) running times for LTM learning using the LTAB algorithm (Wang et al., 2008), b) approximation accuracy of probabilistic inference and c) running time for inference. Approximation accuracy is measured by the Kullback-Leibler divergence between approximate inferred distributions and exact inferred distributions obtained from clique tree propagation on the original BN. $N$ and $C$ designate the sample size and the maximal cardinality of latent variables, respectively. These results come from the work of Wang et al. (2008).

Probabilistic inference is used in many circumstances, such as in credit card fraud detection (Ezawa & Norton, 1996) or disease diagnostic (McKendrick, Gettinbya, Gua, Reidb, & Revie, 2000).

Probabilistic inference in a general BN is known to be an NP-hard task (Cooper, 1990). To tackle this issue, one can approximate the original BN using a maximum weight spanning tree learned relying on Chow and Liu's algorithm (1968). The drawback is the risk of inaccuracy in inference results. In this context, LTM provides an efficient and simple solution, because: (i) thanks to its tree structure, the model allows linear computations with respect to the number of OVs, and at the same time, (ii) it can represent complex relations between OVs through multiple LVs. Because learning LTM before performing inference can be time-consuming Wang et al. (2008) propose the following strategy: first, offline model learning is performed, then answers to probabilistic queries are quickly computed online. However, recent spectral methods (Parikh et al., 2011) considerably reduced model learning phase, because they do not require to learn the model structure. This makes inference through large LTMs possible (around several hundred OVs) and thus renders them even more attractive.

Inferential complexity does not only depend on the number of OVs, but also on LV cardinalities: the higher the cardinality, the higher the complexity. Hence Wang et al. (2008) propose a tradeoff between inferential complexity and model approximation accuracy by fixing a maximal cardinality $C$ for LVs.

Wang et al. (2008) empirically demonstrated the high performance of LTM-based inference on 8 well-known Bayesian networks from the literature. The principle consists in learning the LTM which will provide the best approximation of the original BN. For this purpose, data are sampled from the original BN and then an LTM is learned from the data. To illustrate inference, let us only consider two examples, the ALARM and the BARLEY networks which show the lowest and highest inferential complexities among the aforementioned networks, respectively. The ALARM network contains 37 nodes, and is characterized by an average indegree of 1.24 (max: 4) and an average cardinality of 2.84 (max: 4). The BARLEY network contains 48 nodes; its average indegree is 1.75 (max: 4) and its average cardinality is 8.77 (max: 67). Two parameters are central for the user: (i) $N$, the sample size which entails long model learning running times but leads to better model accuracies, and (ii) $C$, the maximal cardinality of LVs which entails long model learning and inference running times but leads to higher model accuracies.

In Figure 12, the LTM-based method is compared to other standard inference approaches: the LCM-based approach, the Chow-Liu (CL) model-based approach and loopy belief propagation (LBP) (Pearl, 1988). Exact inference through clique tree propagation (CTP) (Lauritzen & Spiegelhalter, 1988) on the original BN is considered as the reference. Figure 12a reports running times for LTM learning using the LTAB algorithm (Wang et al., 2008). Running times almost linearly increase with $N$ and $C$. Regarding inference accuracy (Figure 12b), the LTM-based method outperforms other methods when $N$ and $C$ are high, e.g. $N = 100k$ and $C = 8$ for the ALARM network. As regards inference running times (Figure 12c), the benefits of using the LTM-based method are high for the ALARM network, a high inferential complexity network. In these experiments, we note that CL is also very interesting, and the choice between the latter and the LTM will depend on the online inference time allowed. If time is very limited, CL would be preferred. In the other case, the LTM would be chosen.

### 4.4 Other Applications

Beside latent structure discovery, multidimensional clustering and probabilistic inference, there are other interesting applications of LTMs.

A simple but efficient classifier is naive Bayes. This model assumes that OVs are independent conditional on the class variable. This assumption is often violated by data and hence numerous adaptations have been developed to improve the classifier performance. Naive Bayes has been generalized by introducing latent nodes as internal discrete nodes (Zhang et al., 2004) or continuous nodes (Langseth & Nielsen, 2009), mediating the relation between leaves and the class variable. The model is identical to an LTM except that the root is observed. Recently, Wang et al. (2011) proposed a classifier based on LTM. For each class, a specific LTM is learned and a latent tree classifier is built by aggregating all LTMs. This classifier outperforms naive Bayes and many other successful classifiers such as tree augmented naive Bayes (Friedman, Geiger, & Goldszmidt, 1997) and averaged one-dependence estimator (Webb, Boughton, & Wang, 2005).

More specifically to some research fields, LTM has been used for human interaction recognition, haplotype inference in genetics and diagnosis in traditional medicine. Human interaction recognition is a challenging task, because of multiple body parts and concomitant inclusions (Aggarwal & Cai, 1999). For this purpose, the use of LTM allows to segment the interaction in a multi-level fashion (Park & Aggarwal, 2003): body part positions are estimated through low-level LVs, while overall body position is estimated by a high-level LV. In genetics, there is a need for inferring haplotypic data (*i.e.* latent genetic DNA sequences) from genotypic data (observed data). Kimmel and Shamir (2005) perform efficient haplotypic inference using a two-layer LFM. Finally, Zhang et al. (2008) applied LTMs to traditional Chinese medicine (TCM). They discovered that the latent structure obtained matches TCM theories. The model provides an objective justification for the ancient theories.

## 5. Discussion

In data analysis, LTM represents an emerging popular topic as it offers several advantages:

- The model allows to discover interpretable latent structure.

- Each latent variable is intended to represent a way to cluster categorical data, and connections between latent variables are meant to express relations between the multiple clustering ways.

- Multiple latent variables organized into a tree greatly improve the flexibility of probabilistic modeling while, at the same time, ensuring linear - thus fast - probabilistic inference.

Applications of LTMs are summarized in Table 6, which recapitulates three types of applications with details, examples, references to generic algorithms, scalability to large datasets, software and bibliographical references.

In the past decade, extensive research efforts have been done in LTM learning. When structure is known, standard EM and LCM-based EM or spectral methods can be used.

| Application | Detail | Example | Algorithm | Scalability | Software | Reference |
|---|---|---|---|---|---|---|
| Latent structure discovery | Discovers latent information underlying data and latent relationships existing between observed and latent information, and also between pieces of latent information. | Phylogenetics: Reconstruction of evolutionary relationships between species | NJ | Yes | NINJA | Saitou and Nei, 1987 |
| Multidimensional clustering | Reveals multiple facets of data and partitions data along each facet. Reveals general relations between multiple facets. | Marketing research: Understanding of consumer behavior | AGS | No | EAST | Chen et al., 2011 |
| Probabilistic inference | Allows exact and linear computations for answering probabilistic queries of the form p(x|y), for a event x given some knowledge y. | Medicine: Diagnostic of disease | AHCB | No | LTAB | Wang et al., 2008 |

Table 6: Summary for main applications of the latent tree model.

When structure is unknown, three classes of methods have been proposed: search-based, variable clustering-based and distance-based methods. The first one is slow but leads to accurate models. The second one drastically decreases running times. Finally, the last class guarantees to exactly recover the generative LTM structure under the assumption that all LVs have the number of states and this number is known.

In spite of the aforementioned advances, the use of LTM presents some drawbacks. For example, when the data dimension is large or very large, model learning still remains prohibitive. Regarding probabilistic inference, LTM provides better results than the standard Chow-Liu's approach, but leads to more computational burden.

## 6. Future Directions

Progress on LTM has been made, but there is still much to be done. There are multiple promising directions. For instance, a recent work developed LTM for continuous data analysis (Poon, Zhang, Chen, & Wang, 2010; Choi et al., 2011). Other authors investigated the relationships between LTM and ontology (Hwang et al., 2006), and LTM-based dependence visualization (Mourad, Sinoquet, Dina, & Leray, 2011). Although no research has been carried out on the application to causal discovery and latent trait analysis, we argue that LTM might represent interesting avenues of research.

**LTM for Continuous Data:** Recently, LTM modeling has been applied to continuous data analysis (Poon et al., 2010; Choi et al., 2011; Song et al., 2011; Kirshner, 2012). For instance, Poon et al. (2010) proposed a new model, called pouch latent tree model (PLTM). PLTM is identical to LTM, except that each observed node in an LTM is replaced with a "pouch" node representing an aggregation of multiple continuous OVs. PLTM represents a generalization of the Gaussian mixture model (GMM) when more than one LV is allowed. The proposal of Poon et al. originates from the fact that model-based clustering of continuous data is sensitive to the selected variables. Similarly to categorical clustering (Section 4.2), in high-dimensional continuous data, there are multiple ways to partition the data, and the multiple LVs of PLTM are able to take this into account. Poon et al.

developed a search-based algorithm for PLTM learning. The latter algorithm is closely related to the EAST algorithm (Zhang & Kocka, 2004b) dedicated to LTM learning and is thus quite slow. Hence, the development of new methods dedicated to efficient PLTM learning certainly represents interesting perspectives of research. Besides, a next step would also be the development of LTM dedicated to mixed data analysis, *i.e.* combining categorical and continuous data.

**LTM Structure and Ontology:** It is possible to relate LTM to ontology, in particular taxonomy (tree-structured ontology). For instance, when applying LTM to a microarray dataset of yeast cell-cycle, Hwang et al. (2006) showed that some LVs are significantly related to specific gene ontology terms, such as organelle organization or cellular physiological process. Thus taxonomy could help interpret LVs. Moreover the taxonomy structure could be used as *a priori* structure.

**Dependence Visualization:** LTM provide a compact and interpretable view of dependences between variables, thanks to its graphical nature and its latent variables (Mourad et al., 2011). Compared to heat map (Barrett, Fry, Maller, & Daly, 2005) which can only display pairwise dependences between variables, LTM helps visualize both pairwise and higher-order dependences. Pairwise dependence can be represented by the chain length linking two leaf variables, whereas higher-order dependences are simply represented by a set of leaf variables connected to a common LV.

**Causal Discovery:** We argue that LTM represents simple but efficient model for causal discovery, for the following reasons:

- If the LTM root is known, then the model can be interpreted as a hierarchy. Into this hierarchy, LVs are distributed into multiple layers. This multiple LV layers represent different degrees of information compactness (*i.e.* data dimension reduction), since each LV captures the patterns of its child variables. The connexion of variables through parent-child relationships allows easy and natural moves from general (top layers) to specific (bottom layers) causes, and *vice-versa*. Thus, causal discovery can be guided by the hierarchical model feature.

- After constructing the model on variables $X = \{X_1, ..., X_n\}$, if one wants to test the direct dependence between $X_i$ and another variable $Y$ not present in the network, it can be easily computed through a test for independence between $X_i$ and $Y$ conditional on the parent of $X_i$. A practical advantage of this conditional test meant to assess direct dependence is that the number of degrees of freedom required is low (because only the parent of $X_i$ is used to condition the test), which ensures a good power.

**Latent Trait Analysis:** Similarly to the generalization of LCM by LTM, it would be worth developing an extension of the latent trait model by a tree-structured model where internal nodes are continuous LVs. For instance, this would alleviate the drawbacks of local independence in the latent trait model and would provide multiple facets thanks to LVs when dealing with high-dimensional data.

## Acknowledgments

The authors are deeply indebted to four anonymous reviewers for their invaluable comments and for helping to improve the manuscript. This work was supported by the BIL Bioinformatics Research Project of Pays de la Loire Region, France. The authors are also grateful to Carsten Poulsen (Aalborg University, Denmark) for providing the Danish beer data, Yi Wang (National University of Singapore) for the LTAB algorithm, Tao Chen (EMC Corporation, Beijing, China) for the EAST algorithm, Stefan Harmeling (Max Planck Institute, Germany) for the BIN-A and BIN-G algorithms, and Myung Jin Choi (Two Sigma Investments, USA) and Vincent Tan (University of Wisconsin-Madison, USA) for the RG, CLRG and regCLRG algorithms.

## Appendix A. Online Resources Mentioned

Software:

- **BIN-A, BIN-G, CL and LCM:**
  http://people.kyb.tuebingen.mpg.de/harmeling/code/ltt-1.4.tar

- **CFHLC:**
  https://sites.google.com/site/raphaelmouradeng/home/programs

- **DHC, SHC and HSHC:**
  http://www.cse.ust.hk/faculty/lzhang/ltm/index.htm (hlcm-distribute.zip)
  http://www.cse.ust.hk/faculty/lzhang/ltm/index.htm (toolBox.zip)

- **EAST:**
  http://www.cse.ust.hk/faculty/lzhang/ltm/index.htm (EAST.zip)

- **Lantern:**
  http://www.cse.ust.hk/faculty/lzhang/ltm/index.htm (Lantern2.0-beta.exe)

- **NJ, RG, CLRG and regCLRG:**
  http://people.csail.mit.edu/myungjin/latentTree.html

- **NJ (fast implementation):**
  http://nimbletwist.com/software/ninja

All datasets used for the algorithm comparison are available at :
https://sites.google.com/site/raphaelmouradeng/home/programs

## Appendix B. Supplemental Material

### B.1 Experiments on Parameter Learning with EM

We studied the number of random restarts required, in practice, to obtain the convergence of EM (Section 3.1.1) and LCMB-EM (Section 3.1.2) to the optimal solution. BIC scores are presented in Figure 13. For EM, we used the method of Chickering and Heckerman (1997) which is implemented in the software LTAB (Wang et al., 2008). We analyzed three datasets: two small ones (BinTree and Asia) and a large one (Tree). For the first two, convergence was achieved after 20 and 2000 restarts, respectively. For the large dataset, convergence is never achieved, even after 5000 restarts. For LCMB-EM, we used the software BIN-A (Harmeling & Williams, 2011). We analyzed three datasets: one small one (Asia) and two large ones (Tree, Alarm). Convergence is achieved for the first two datasets with only one parameter initialization, whereas for the later which is the largest, convergence is never achieved (we were not able to assess with more than 1000 restarts because of a prohibitive running time).

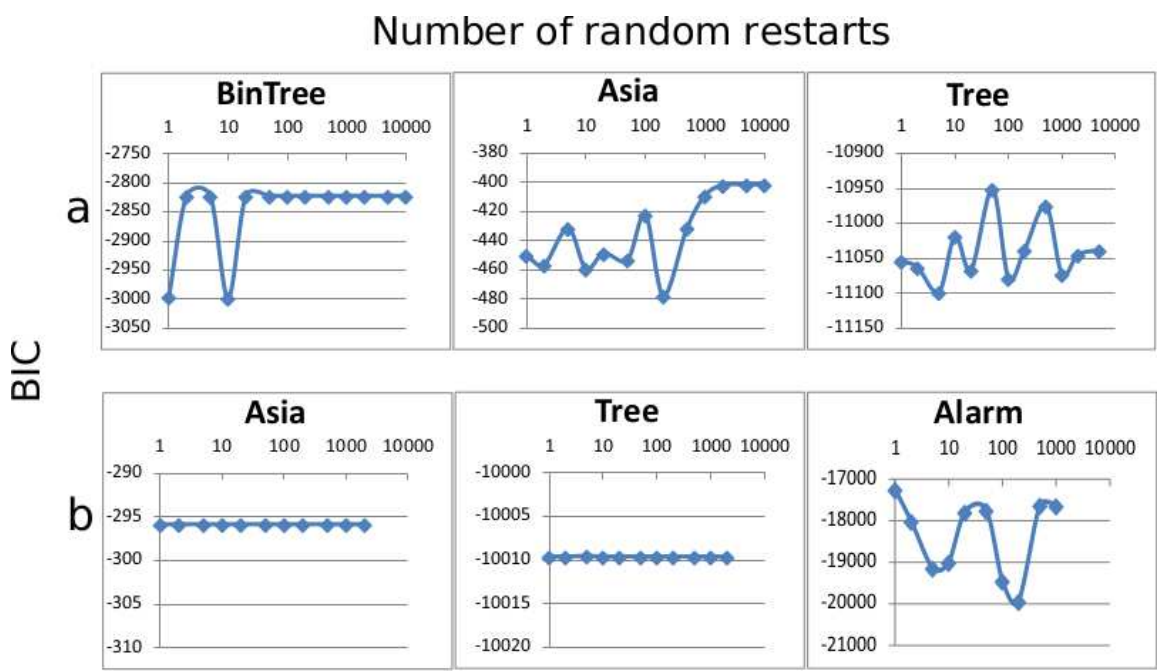

Figure 13: The impact of the number of random restarts on the convergence of expectation-maximization (EM). a) EM. b) LCMB-EM. The number of restarts is reported on the x-axis while the y-axis indicates BIC scores.

### B.2 Time Complexity

We recall the reader that $n$ is the number of variables (input data), $N$ the number of observations and $s$ the number of steps (for search-based algorithms). The time complexities of generic algorithms for LTM learning are detailed as follows:

- **Naive greedy search (NGS)**. There are $O(s)$ steps needed for the convergence of search-based methods. For each step, there are $O(n^2)$ new models generated through the use of 3 operators: addition/removal of an LV and node relocation (Zhang, 2004). For each model, the cardinality is optimized for each LV, so that $O(n^2)$ new models are generated (Zhang, 2004). For each model, parameters are learned using EM, which is achieved in $O(nN)$. Thus, the overall complexity is : $O(s) * O(n^2) * O(n^2) * O(nN) = O(sn^5N)$.

- **Advanced greedy search (AGS), Algorithm 2**. There are $O(s)$ steps needed for the convergence of search-based methods. For each step, there are $O(n^2)$ new models generated through the use of 5 operators: addition/removal of an LV, node relocation and addition/dismissal of a state relative to an LV (Zhang & Kocka, 2004b). For each model, model evaluation is realized through local EM in $O(N)$. After choosing the best model at each step, parameters are learned using EM, which is achieved in $O(nN)$. Thus, the overall complexity is : $O(s) * (O(n^2) * O(N) + O(nN)) = O(sn^2N)$.

- **Agglomerative hierarchical clustering-based learning (AHCB)**. The agglomerative hierarchical clustering is achieved in $O(n^2N)$[19]. LV cardinality and parameters can be learned in $O(N)$ thanks to LCM parameter learning. There are $O(n)$ LVs, thus the complexity is $(O(N) * O(n))$. A final global EM parameter learning is done in $O(nN)$. Thus, the overall complexity is : $O(n^2N) + (O(N) * O(n)) + O(nN) = O(n^2N)$.

- **Latent class model-based LTM learning (LCMB-LTM), Algorithm 3**. Pairwise mutual information values are computed in $O(n^2N)$. LCM is learned in $O(N)$. LCM learning is called $O(n)$ times, *i.e.* for each new LV added to the model. LV cardinality and parameters are learned during LCM learning. A final global EM parameter learning is done in $O(nN)$. Thus, the overall complexity is : $O(n^2N) + (O(N) * O(n)) + O(nN) = O(n^2N)$.

- **Neighbor joining (NJ)**. To learn the structure, there are $O(n)$ steps. At each step, pairwise distances are computed in $O(n^2N)$. Structure learning thus requires $O(n^3N)$ computations. After learning the structure, parameters can be learned with EM or LCMB-EM in $O(nN)$. Thus, the overall complexity is : $O(n^3N) + O(nN) = O(n^3N)$.

- **Distance-based general LTM learning (DBG), Algorithm 4**. First the distances are computed in $O(n^3N)$. If the minimum spanning tree is learned before, the complexity is reduced to $O(n^2N)$. Then, to learn the structure, testing child-parent and sibling relations necessitates $O(n^4)$ operations in the worst case, *i.e.* when the tree is a hidden Markov model. Parameters can be learned with EM or LCMB-EM in $O(nN)$. Thus, the overall complexity is : $O(n^3N) + O(n^4) + O(nN) = O(n^3N + n^4)$ or $O(n^2N) + O(n^4) + O(nN) = O(n^2N + n^4)$.

### B.3 Description of Datasets Used for Literature Algorithm Comparison

Small datasets ($n \leq 10$ variables):

- **BinTree.** Datasets generated using a binary tree on 7 variables (4 leaves and 3 internal nodes), each having eight states. Only leaf data are used. The train and test datasets both consist of 500 observations. The model comes from the work of Harmeling and Williams (2011).

---

[19]. The complexity of agglomerative hierarchical clustering is $O(n^2N)$ using the single linkage criterion. The complexity is higher for other criteria.

- **BinForest.** Datasets generated from a binary forest composed of two trees. One tree has 3 variables (2 leaves and 1 internal node), the other one has 5 variables (3 leaves and 2 internal nodes). Only leaf data are used. The train and test datasets both consist of 500 observations. The model comes from the work of Harmeling and Williams (2011).

- **Asia.** Datasets generated using the well-known Asia network containing 8 binary OVs. The train and test datasets both consist of 100 observations.

- **Hannover.** Real dataset containing 5 binary variables. The dataset has been split into a train dataset and a test dataset. They consist of 3573 and 3589 observations, respectively. The dataset comes from the work of Zhang (2004).

- **Car.** Real dataset containing 7 variables. The dataset has been split into a train dataset and a test dataset. They consist of 859 and 869 observations, respectively. The dataset is available at :
http://archive.ics.uci.edu/ml/.

Large datasets ($10 \leq n \leq 100$ variables):

- **Tree.** Datasets generated using a tree on 50 variables (19 leaves and 31 internal nodes). Only leaf data are used. The train and test datasets both consist of 500 observations.

- **Forest.** Datasets generated using a tree on 50 variables (20 leaves and 30 internal nodes). Only leaf data are used. The train and test datasets both consist of 500 observations.

- **Alarm.** Datasets generated using the well-known Alarm network containing 37 OVs. The train and test datasets both consist of 1000 observations.

- **Coil-42.** Real dataset containing 42 variables. The dataset has been split into a train dataset and a test dataset. They consist of 5822 and 4000 observations, respectively. The dataset comes from the work of Zhang and Kocka (2004b).

- **NewsGroup.** Real dataset containing 100 binary variables. The dataset has been split into a train dataset and a test dataset. They both consist of 8121 observations. The dataset is available at :
http://cs.nyu.edu/roweis/data/20news_w100.mat.

Very large datasets ($n > 100$ variables):

- **HapGen.** Datasets generated over 1000 genetic variables using the HAPGEN software (Spencer, Su, Donnelly, & Marchini, 2009). The train and test datasets both consist of 1000 observations.

- **HapMap.** Real dataset containing 10000 binary variables. The dataset has been split into a train dataset and a test dataset. They consist of 118 and 116 observations, respectively. The dataset comes from HapMap phase III (The International HapMap Consortium, 2007) and concerns Utah residents with Northern and Western European ancestry (CEU).